\documentclass{article} 

\usepackage{amsmath}
\usepackage{comment}

\usepackage[preprint]{neurips_2026}

\usepackage[utf8]{inputenc} 
\usepackage[T1]{fontenc}    
\usepackage{hyperref}       
\usepackage{url}            
\usepackage{booktabs}       
\usepackage{amsfonts}       
\usepackage{nicefrac}       
\usepackage{microtype}      
\usepackage{xcolor}         

\usepackage{tikz}           
\usepackage{pifont}
\newcommand{\cmark}{\ding{51}} 
\newcommand{\xmark}{\ding{55}} 
\usetikzlibrary{arrows.meta, positioning, shapes, fit, calc, backgrounds} 
\usepackage{multirow} 
\usepackage{placeins}
\usepackage{rotating}
\usepackage{algorithm}
\usepackage{algpseudocode}

\usepackage{tabularx}
\usepackage{amsmath}
\usepackage{array}
\usepackage{multirow}

\usepackage{enumitem}

\title{Joint Treatment Effect Estimation from Incomplete Healthcare Data: Temporal Causal Normalizing Flows with LLM-driven Evolutionary MNAR Imputation}

\author{%
  \normalfont
  \begin{tabular}{cc}
    \begin{tabular}[t]{c}
      {\bfseries Olivia Jullian Parra}\thanks{Equal contribution.} \\
      Department of Mathematical Modeling\\
      and Machine Learning\\
      University of Zürich\\
      Winterthurerstrasse 190, Zürich 8057\\
      \texttt{olivia.jullianparra@uzh.ch}
    \end{tabular}
    &
    \begin{tabular}[t]{c}
      {\bfseries Sara Zoccheddu}\footnotemark[1] \\
      Department of Mathematical Modeling\\
      and Machine Learning\\
      University of Zürich\\
      Winterthurerstrasse 190, Zürich 8057\\
      \texttt{sara.zoccheddu@uzh.ch}
    \end{tabular}
    \\[7em]
    \begin{tabular}[t]{c}
      {\bfseries David Catalan Cerezo} \\
      Department of Mathematics\\
      ETH Zürich\\
      Rämistrasse 101, Zürich 8092\\
      \texttt{dcatalan@ethz.ch}
    \end{tabular}
    &
    \begin{tabular}[t]{c}
      {\bfseries Tom Forzy} \\
      Epidemiology, Biostatistics \& Prevention\\
      Institute (EBPI), University of Zürich\\
      Hirschengraben 84, Zürich 8001\\
      \texttt{tom.forzy@uzh.ch}
    \end{tabular}
    \\[6em]
    \begin{tabular}[t]{c}
      {\bfseries Franziska Ulrich} \\
      Department of Mathematics\\
      ETH Zürich\\
      Rämistrasse 101, Zürich 8092\\
      \texttt{ulricfr@ethz.ch}
    \end{tabular}
    &
    \begin{tabular}[t]{c}
      {\bfseries William Sutcliffe} \\
      Department of Physics\\
      University of Zürich\\
      Winterthurerstrasse 190, Zürich 8057\\
      \texttt{william.sutcliffe@physik.uzh.ch}
    \end{tabular}
    \\[6em]
    \begin{tabular}[t]{c}
      {\bfseries Jakob Martin Burgstaller} \\
      Institute of Primary Care\\
      University of Zürich\\
      Sonneggstrasse 6, Zürich 8091\\
      \texttt{jakobmartin.burgstaller@usz.ch}
    \end{tabular}
    &
    \begin{tabular}[t]{c}
      {\bfseries Oliver Senn} \\
      Institute of Primary Care\\
      University of Zürich\\
      Sonneggstrasse 6, Zürich 8091\\
      \texttt{oliver.senn@usz.ch}
    \end{tabular}
    \\[6em]
    \begin{tabular}[t]{c}
      {\bfseries Patrick Owen} \\
      Department of Physics\\
      University of Zürich\\
      Winterthurerstrasse 190, Zürich 8057\\
      \texttt{powen@physik.uzh.ch}
    \end{tabular}
    &
    \begin{tabular}[t]{c}
      {\bfseries Nicola Serra} \\
      Department of Physics \&\\
      Department of Mathematical Modeling\\
      and Machine Learning\\
      University of Zürich\\
      Winterthurerstrasse 190, Zürich 8057\\
      \texttt{nicola.serra@physik.uzh.ch}
    \end{tabular}
  \end{tabular}
}

\begin{document}

\maketitle

\begin{abstract}

Target trial emulation (TTE) provides a framework for answering causal questions using observational data when randomized controlled trials (RCTs) are infeasible. However, standard methods for treatment effect estimation have been developed in isolation, failing to jointly address the compounding challenges inherent to analyses of observational data such as electronic health records (EHRs). In particular, these challenges include time-varying confounding and missing-not-at-random (MNAR) missingness reaching 50\%--80\% for critical biomarkers. To address this gap, we propose a two-stage pipeline that jointly handles MNAR missingness and causal structure across time. CausalFlow-T, a Directed Acyclic Graph (DAG)-constrained normalizing flow with Long Short-Term Memory (LSTM)-encoded patient history, performs exact invertible counterfactual inference, eliminating the approximation errors and confounding biases where existing variational and adversarial methods fail silently. Ablations on four synthetic and one semi-synthetic dataset with known counterfactuals confirm that its two core design choices address strictly non-overlapping failure modes (DAG constraints for confounding separation, exact inference for structural propagation) with neither compensating for the absence of the other. To handle the incomplete data CausalFlow-T receives as input, we propose an LLM-driven evolutionary imputer and evaluate it with three LLM backends, including two open-source models. Across 30\%--80\% MNAR missingness, the imputer achieves the best pooled rank across biomarker and causal metrics, leading on point-wise accuracy and temporal extrapolation while maintaining average treatment effect (ATE) recovery where statistical methods progressively degrade. Applied to a cohort of adults with type 2 diabetes in 
Swiss primary care initiating a GLP-1 receptor agonist 
or SGLT-2 inhibitor, the pipeline recovers a 
per-protocol weight-loss difference of $-0.98$ kg [$95\%$ CI $-1.01$, $-0.96$] favoring GLP-1 receptor agonists, consistent with 
RCT evidence and estimated directly from realistically 
incomplete real-world EHR data.

\end{abstract}

\section{Introduction}
\label{sec:intro}
Type 2 diabetes (T2D) is one of the most common chronic conditions affecting over half a billion adults worldwide~\citep{InternationalDiabetesFederation2025IDFEdition}. To optimize glycemic 
control and reduce the risk of complications (e.g. cardiovascular 
disease), effective 
management requires a multifactorial approach~\citep{Committee202510.Diabetes2025}. Recent therapeutic advances, particularly two drug classes, GLP-1 receptor agonists (GLP-1RAs; including semaglutide, 
marketed as Ozempic) and 
sodium-glucose co-transporter-2 inhibitors (SGLT-2is), have transformed T2D management by demonstrating cardiovascular, renal, and weight loss benefits beyond glycemic control in randomized controlled trials (RCTs)~\citep{Committee202510.Diabetes2025}. However, RCTs are limited by high costs, restricted sample sizes, and strict eligibility criteria, making them insufficient to address the complexity of T2D treatment, where therapies are numerous, combined, and evolve over time~\citep{Schneeweiss2021ConductingTreatments}. Observational data (e.g., electronic health records [EHRs]) can 
complement RCTs by providing real-world evidence across diverse patient 
populations and care settings~\citep{Schneeweiss2021ConductingTreatments}. Yet, causal inference from EHRs remains challenging due to confounding, selection bias, and other structural biases arising from non-randomized treatment assignment. Target trial emulation (TTE) addresses this by explicitly framing observational analyses to mimic a hypothetical RCT, thereby improving causal validity~\citep{Hernan2016PracticeAvailable}. Despite this framework, existing methods for treatment effect estimation have been developed in isolation, and none jointly addresses the structural challenges of real EHR data~\citep{shalit2017,louizos2017,bica2020,d2021overlap}. First, treatment decisions are driven by the patient's evolving clinical profile, creating time-varying confounding that existing methods handle only under parametric assumptions or without explicit causal structure~\citep{shalit2017, louizos2017, bica2020, d2021overlap}. Second, EHR data exhibit substantial missingness (often exceeding $>50\%$) that arises under missing-not-at-random (MNAR) mechanisms (clinical decisions drive data collection) while imputation strategies have been typically validated at far lower missingness rates~\citep{curnow2024multiple,he2025llm,mangussi2026large}. These challenges interact: MNAR missingness distorts biomarker (e.g., blood pressure)–outcome relationships, compounding bias in downstream causal estimation.

We address this gap with a two-stage pipeline that jointly handles time-varying confounding and MNAR missingness within the TTE framework. We apply this pipeline to estimate the per-protocol effect of GLP-1RAs versus SGLT-2is on 1-year body weight in a Swiss primary care cohort of adults with T2D, extending existing RCT evidence from specific agents in selected populations to the full drug classes used in real-world clinical practice. Our contributions are as follows:

\textbf{CausalFlow-T:} A directed acyclic graph (DAG)-constrained normalizing flow (NF) conditioning a causal masked autoregressive flow (CausalMAF) on a LSTM-encoded patient history to model the joint distribution of covariates, treatment, and outcomes over time. Experiments on four 
    synthetic datasets and one semi-synthetic evaluation dataset with known 
    counterfactuals reveal two distinct failure modes:
    (i)~DAG constraints are necessary for confounding separation (unconstrained models maintain systematic bias in low-effect subgroups
    even despite low factual prediction error); (ii)~exact inference is required for structural propagation (variational methods fail to recover
    the true treatment effect under complex mediation paths even with correctly specified causal graphs). 
    
\textbf{LLM-driven evolutionary imputation:} A large language model (LLM)-driven
    evolutionary pipeline benchmarked against LOCF~\citep{fitzmaurice2012applied}, MissForest~\citep{stekhoven2012missforest}, and a DAG-aware flow matching
    (CFM) baseline (inspired by the m-graph framework~\citep{mohan2021graphical})
    across 30--80\% MNAR missingness using three LLM backends (GPT-5.4 and GPT-OSS-120b from OpenAI,
    Qwen3.5-Plus from Alibaba Cloud). Evaluated via reconstruction, causal correlation structure preservation, and downstream ATE recovery, LLM-based methods dominate, with GPT-5.4 achieving the best overall performance and open-source models confirming robustness and reconstruction–causal trade-offs.

\textbf{Real-world application:} When applied to a Swiss primary care T2D cohort initiating GLP-1RAs or SGLT-2is, our two-stage pipeline produces robust per-protocol estimates of 1-year body weight change (kg) from incomplete EHR data.


\section{Related Work}
\label{sec:related}

\textbf{Treatment effect estimation and time-varying confounding:}
Marginal structural models (MSMs) with inverse probability weighting 
(IPW)~\citep{hernan2020} address time-varying confounding but rely on parametric assumptions that 
break down in high-dimensional, nonlinear 
settings~\citep{d2021overlap}. Deep learning (DL) approaches such as 
R-MSN~\citep{lim2018}, CRN~\citep{bica2020}, and the Causal 
Transformer~\citep{melnychuk2022} improve flexibility but rely on 
approximate inference or lack explicit causal structure, making them 
prone to spurious associations under interventions~\citep{javaloy2023}. Generative models such as
CEVAE~\citep{louizos2017} and GANITE~\citep{yoon2018} introduce
additional approximation gaps via variational or adversarial objectives.
All have been primarily evaluated on synthetic datasets with limited
missingness and large sample sizes~\citep{shalit2017}, conditions 
rarely met in real-world clinical 
data~\citep{sun2024incorporating,barrett2020selective}.

\textbf{Normalizing flows for causal inference:}
~\citep{khemakhem2021} establish identifiability via
nonlinear independent component analysis for autoregressive flows; ~\citep{javaloy2023}
extend this to general triangular mappings with a full do-operator
formulation. ~\citep{chao2023modeling} use diffusion
models at the node level, but requiring a separate network per variable
limits scalability for high-dimensional EHR data. All existing methods assume static data. We extend DAG-constrained flows to temporal settings by conditioning on LSTM-encoded patient histories, capturing distributional shifts over time in a unified model for longitudinal treatment effect estimation.

\textbf{Missing data handling in causal inference:}
Missing data in EHRs can bias treatment effect estimation even after imputation~\citep{zhou2023}. Effective imputation must leverage temporal disease trajectories and account for the underlying missingness mechanism~\citep{zhou2023}.
Standard methods such as Multiple Imputation by Chained Equations (MICE)~\citep{sterne2009} and
missing forest (MissForest)~\citep{stekhoven2012missforest} assume missing-at-random (MAR) and can amplify bias under MNAR, while 
LOCF distorts temporal correlation structure by construction. 
Recent LLM-based approaches show improved reconstruction, but have been
validated mainly at regimes of
5--40\% missingness~\citep{he2025llm,mangussi2026large}, below the high-rate MNAR
missingness often encountered in real-world EHR data;
in parallel, LLMs have been used as mutation operators in evolutionary
program search for mathematical and algorithmic discovery~\citep{
romeraparedes2024mathematical,liu2024evolution,novikov2025alphaevolve,
lange2025shinkaevolve}.
\section{Methodology}
\label{sec:overview}

We propose a two-stage pipeline for treatment effect estimation 
from incomplete longitudinal observational data. CausalFlow-T (Section~\ref{sec:model}) addresses time-varying confounding and exact counterfactual inference, while the LLM-driven evolutionary imputer (Section~\ref{sec:llm}) handles MNAR missingness and feeds completed data into CausalFlow-T. Validating the pipeline sequentially isolates the contribution of each stage, ensuring that downstream performance differences reflect imputation quality rather than estimator failure.

\subsection{Problem formulation and assumptions}
\label{sec:overview:problem}

Consider $N$ patients over $T$ discrete time steps. At 
each time $t$, patient $i$ has covariates
$x_{i,t}\in\mathbb{R}^d$ (indexed by $j\in\{1,\ldots,d\}$), treatment $a_{i,t}\in\{0,1\}$, outcome 
$y_{i,t}\in\mathbb{R}$, and missingness mask 
$m_{i,t}\in\{0,1\}^d$. Causal relationships between $(x,a,y)$ 
are encoded in an expert-specified DAG $\mathcal{G}=(V,E)$. The 
incomplete longitudinal dataset is \[ \widetilde{\mathcal{D}} = \{(\tilde{x}_{i,t}, a_{i,t}, y_{i,t}, m_{i,t})\}_{i=1,t=1}^{N,T}\]
 where entry $(i,t,j)$ is observed if and only if $m_{i,t,j}=0$. 
Missingness is MNAR: the probability of a 
value being unobserved depends on the unobserved value itself,
\begin{equation}
  P(m_{i,t,j}=1 \mid x_{i,t,j},\, x_{i,t,-j},\, a_{i,t},\, y_{i,t})
  = f_j\bigl(x_{i,t,j},\, x_{i,t,-j},\, a_{i,t},\, y_{i,t}\bigr),
  \label{eq:mnar}
\end{equation}
where $f_j$ depends on $x_{i,t,j}$ itself
(the MNAR component), as well as on other observed covariates
$x_{i,t,-j}$ and on the outcome state $y_{i,t}$, mirroring the
clinical pattern where tests ordered on suspicion produce
structured, not random, absence. Our goal is to estimate the $\mathrm{ATE} = \mathbb{E}\bigl[y_{i,t}(1) - y_{i,t}(0)\bigr]$, where $y_{i,t}(a')$ denotes the potential outcome under 
$\mathrm{do}(A=a')$~\citep{pearl2009}, as well as individual 
treatment effects $\tau_{i,t} = y_{i,t}(1) - y_{i,t}(0)$.

\textbf{Assumptions.}
We require (i) \textit{sequential ignorability} $A_t \perp\!\!\!\perp Y_t(a') \;\mid\; h_t,\; \mathcal{G}$, where $h_t$ is the LSTM-encoded patient history serving as the adjustment set; and (ii)
\textit{correct DAG specification}: $\mathcal{G}$ correctly encodes the causal structure of $(x,a,y)$; misspecification propagates systematic bias through the abduction-action-prediction (AAP) procedure regardless of estimator quality (sensitivity analysis in Appendix~\ref{app:dag_sensitivity}).

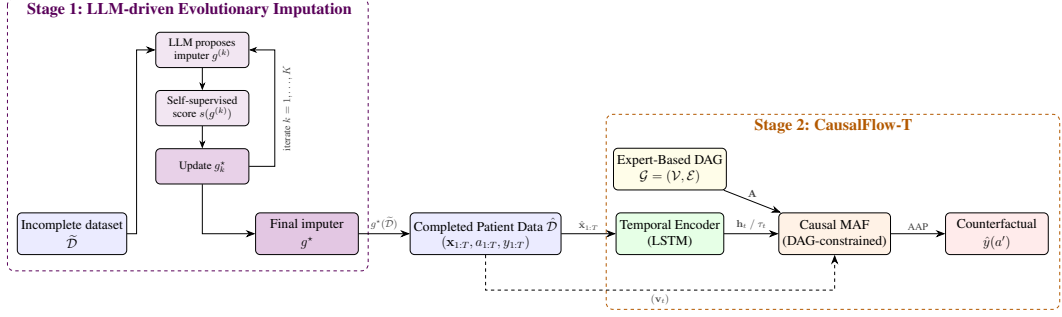
\begin{figure*}[tb]
  \centering
  \resizebox{\textwidth}{!}{%
  \begin{tikzpicture}[
    node distance=1.0cm and 1.5cm,
    block/.style  = {rectangle, draw, rounded corners=4pt,
                     minimum height=1.3cm, minimum width=3.2cm,
                     align=center, font=\large},
    sblock/.style = {rectangle, draw, rounded corners=4pt,
                     minimum height=1.1cm, minimum width=2.9cm,
                     align=center, font=\normalsize},
    arrow/.style    = {-{Stealth[length=3mm]}, thick, black},
    looparrow/.style= {-{Stealth[length=2.5mm]}, thick, black},
    lbl/.style    = {font=\small, text=black!75},
    stagelbl/.style   = {font=\bfseries\Large},
    stageframe/.style = {draw, dashed, rounded corners=6pt, thick}
  ]

  \node[block, fill=blue!8] (raw)
    {Incomplete dataset\\[2pt]$\widetilde{\mathcal{D}}$};

  \node[block, fill=violet!22, right=4.0cm of raw] (finalg)
    {Final imputer\\[2pt]$g^\star$};

  \node[block, fill=blue!8, right=1.6cm of finalg] (done){Completed Patient Data $\hat{\mathcal{D}}$\\$(\mathbf{x}_{1:T}, a_{1:T}, y_{1:T})$};

  \node[block, fill=green!10, right=1.7cm of done] (enc)
    {Temporal Encoder\\(LSTM)};

  \node[block, fill=orange!10, right=1.7cm of enc] (maf)
    {Causal MAF\\(DAG-constrained)};

  \node[block, fill=red!8, right=1.7cm of maf] (out)
    {Counterfactual \\[2pt]$\hat{y}(a')$};

  \coordinate (lcx) at ($(raw)!0.5!(finalg)$);

  \node[sblock, fill=violet!18] (update)
    at ($(lcx)+(0.4,2.1)$) {Update $g^\star_k$};
  \node[sblock, fill=violet!10, above=7mm of update] (score)
    {Self-supervised\\score $s(g^{(k)})$};
  \node[sblock, fill=violet!10, above=7mm of score]  (propose)
    {LLM proposes\\imputer $g^{(k)}$};

  \draw[arrow] (raw.east) -- ++(3mm,0) |- (propose.west);
  \draw[looparrow] (propose) -- (score);
  \draw[looparrow] (score)   -- (update);

  \draw[looparrow] (update.east) -- ++(8mm,0) coordinate (lc1)
                   |- (propose.east);
  \node[lbl, rotate=90]
    at ($(lc1)!0.5!(lc1 |- propose.east)+(4mm,0)$)
    {iterate $k=1,\dots,K$};

    \draw[arrow] (update.south) |- (finalg.west)
      node[pos=0.7, above, lbl] {};

  \draw[arrow] (finalg) -- (done)
    node[midway, above, lbl] {$g^\star(\widetilde{\mathcal{D}})$};
  \draw[arrow] (done) -- (enc)
    node[midway, above, lbl] {$\hat{\mathbf{x}}_{1:T}$};
  \draw[arrow] (enc)  -- (maf)
    node[midway, above, lbl] {$\mathbf{h}_t\;/\;\tau_t$};
  \draw[arrow] (maf)  -- (out)
    node[midway, above, lbl] {AAP};

  \node[block, fill=yellow!10, above=0.7cm of enc] (dag)
    {Expert-Based DAG\\$\mathcal{G}=(\mathcal{V},\mathcal{E})$};
  \draw[arrow] (dag) -- (maf)
    node[midway, above, lbl] {$\mathbf{A}$};

  \draw[arrow, dashed] (done.south) -- ++(0,-1.1) -| (maf.south)
    node[near start, below, lbl] {$(\mathbf{v}_t)$};

  \begin{scope}[on background layer]
    \draw[stageframe, draw=violet!70!black]
      ($(raw.north west |- propose.north)+(-0.30, 1.05)$)
      rectangle
      ($(finalg.south east |- raw.south)+(0.30,-0.50)$);

    \draw[stageframe, draw=orange!70!black]
      ($(enc.north west |- dag.north)+(-0.30, 1.05)$)
      rectangle
      ($(out.south east)+(0.35,-1.70)$);
  \end{scope}

  \node[stagelbl, text=violet!70!black]
    at ($(raw.north |- propose.north)!0.5!(finalg.north |- propose.north)+(0,0.70)$)
    {Stage~1:~LLM-driven Evolutionary Imputation};

  \node[stagelbl, text=orange!70!black]
    at ($(enc.north |- dag.north)!0.5!(out.north |- dag.north)+(0,0.70)$)
    {Stage~2:~CausalFlow-T};

  \end{tikzpicture} }
  \caption{
  \textbf{Stage~1 (LLM-driven Evolutionary Imputation):} an LLM iteratively
  proposes candidate imputers $g^{(k)}$, scores them via a
  self-supervised proxy $s(g^{(k)})$, and updates the running best
  $g^\star_k$; after $K$ rounds, $g^\star$ produces $\hat{\mathcal{D}}$.
  \textbf{Stage~2 (CausalFlow-T):} a temporal encoder conditions a
  DAG-constrained Causal MAF on patient history; counterfactual outcomes
  $\hat{y}(a')$ are obtained via AAP.}
  \label{fig:arch}
\end{figure*}

\textbf{Two-stage pipeline.}
We first validate CausalFlow-T on complete-data settings 
$\mathcal{D}$ with known counterfactuals, establishing its 
causal estimation properties independently of any imputation 
choices. Once validated, the LLM-driven evolutionary imputer 
produces  $\hat{\mathcal{D}}$, enabling 
CausalFlow-T to operate on realistic MNAR data with following two-stage pipeline (Figure~\ref{fig:arch}):
\[
\widetilde{\mathcal{D}}
\xrightarrow{g^{\star}}
\hat{\mathcal{D}}
\xrightarrow{\mathrm{CausalFlow\text{-}T}}
\{\hat{y}_{i,t}(a'), \hat{\tau}_{i,t}\}
\]

\subsection{CausalFlow-T for Exact Counterfactual Inference}
\label{sec:model}

CausalFlow-T estimates $\hat{y}_{i,t}(a')$ and 
$\hat{\tau}_{i,t}$ via a DAG-constrained normalizing flow 
conditioned on longitudinal patient history, validated first on $\hat{\mathcal{D}}$ where ground-truth 
counterfactuals are known.

\textbf{DAG-Constrained Causal MAF.}
The core generative model is a Causal Masked Autoregressive 
Flow (CausalMAF)~\citep{javaloy2023}. For variable vector 
$v_t = [x_t, a_t, y_t]\in\mathbb{R}^{{d}}$ the flow learns an invertible 
mapping $f_\theta : v_t \mapsto z_t$ with 
$z_t \sim \mathcal{N}(0,I)$, with autoregressive structure 
constrained to a topological sort of $\mathcal{G}$:
\begin{equation}
  v_j = f_j^{-1}(z_j;\, v_{\mathrm{pa}(j)},\, h_t), 
  \quad j=1,\ldots,{d},
  \label{eq:causalflow}
\end{equation}
where $\mathrm{pa}(j)$ denotes the causal parents of node $j$ 
in $\mathcal{G}$. Intervening on $a_t$ therefore propagates 
exclusively through causal descendants while non-descendants 
are held fixed, implementing $P(Y\mid\mathrm{do}(a'),X)$ 
directly. Without this constraint, an unconstrained flow can
propagate interventions through arbitrary learned dependencies 
(including anti-causal directions) producing 
systematically biased counterfactuals even when the factual 
distribution is correctly fitted (Section~\ref{sec:exp:causalflowt}).

The flow maximizes the exact log-likelihood $\log p(v_t)=\log p_z(f_\theta(v_t))+\log\bigl|\det \frac{\partial f_\theta}{\partial v_t}\bigr|$
eliminating the ELBO approximation gap that accumulates under 
variational inference. CVAE and GNN-CVAE collapse to 
a hazard ratio ($\mathrm{HR})\approx 1.0$ on CVD Risk despite a true 18\% hazard reduction, 
consistent with posterior collapse under the Kullback-Leibler(KL) regularization (Section~\ref{sec:exp:causalflowt}, Appendix~\ref{app:theory:elbo}).

\textbf{Temporal encoding via LSTM.}
The CausalMAF is conditioned on an LSTM encoder:
\begin{equation}
  h_t = \mathrm{LSTM}(x_t,\, h_{t-1}), 
  \qquad v_t \sim \mathrm{CausalMAF}(\,\cdot\mid h_t),
  \label{eq:lstm}
\end{equation}
The hidden state $h_t\in\mathbb{R}^H$ summarizes the full observed covariate 
history up to time $t$, capturing disease progression, prior 
treatment exposure, and time-varying confounding. The DAG 
$\mathcal{G}$ governs the contemporaneous factorization 
$P(X_t, A_t, Y_t \mid h_t)$, enabling valid do-calculus~\citep{pearl2009}.

\textbf{Counterfactual inference via AAP.}
Given the trained flow, counterfactuals are computed via 
Pearl's AAP procedure~\citep{pearl2009}. \textit{Abduction}: invert the flow exactly to 
    recover the exogenous noise $z_t = f_\theta(v_t;\, h_t)$, 
    capturing all individual-specific factors not explained 
    by observed covariates. \textit{Action}: replace $a_t \leftarrow a'$; because 
    the autoregressive ordering respects $\mathcal{G}$, the 
    intervention propagates only to causal descendants of 
    $a_t$; $z_t$ is held fixed across both arms, implementing 
    the surgical intervention of the do-operator. \textit{Prediction}: propagate through $f_\theta^{-1}$ 
    via the DAG ordering to obtain $v_t^{\mathrm{cf}}$ and 
    $\hat\tau_{i,t} = \hat y_{i,t}(1) - \hat y_{i,t}(0)$. Because flow inversion is exact, the same $z_t$ represents the individual-specific exogenous noise under both treatment arms, implementing the twin-network assumption and enabling individual-level counterfactuals, a guarantee CVAE-based approaches cannot provide (Appendix~\ref{app:theory:elbo}).

\subsection{LLM-driven evolutionary imputation under MNAR}
\label{sec:llm}
This stage constructs $\hat{\mathcal{D}}$ from 
$\widetilde{\mathcal{D}}$ by evolutionary search over 
imputation operators. Crucially, the LLM does not fill missing 
entries directly; it proposes candidate Python
imputers $\{g^{(k)}\}_{k=0}^{K}$, each a complete module $g^{(k)} : \widetilde{\mathcal{D}} \mapsto 
\hat{\mathcal{D}}^{(k)}$ (full algorithm in Algorithm~\ref{alg:llm_search}, Appendix~\ref{app:theory_llm}).

\textbf{Self-supervised proxy score.}
Let $\Omega_{\mathrm{obs}}=\{(i,t,j):m_{i,t,j}=0\}$. We draw a proxy holdout $\Omega_p \subset \Omega_{\text{obs}}$ by
masking each observed cell independently with probability $\rho\in(0,1)$ (contiguous-run masking in real EHR; Appendix~\ref{app:theory_llm}), forming $\widetilde{\mathcal{D}}_{-}$. The
composite proxy score penalizes both pointwise reconstruction error and
distortion of covariate--treatment and covariate--outcome correlations:

\begin{equation}
s(g)
=
\underbrace{
\sqrt{
\frac{1}{|\Omega_p|}
\sum_{(i,t,j)\in\Omega_p}
\left(
\bigl[g(\widetilde{\mathcal{D}}_{-})\bigr]_{i,t,j}
-
\tilde{x}_{i,t,j}
\right)^2
}
}_{\mathrm{RMSE}(g)}
+
\lambda_Y \Delta_Y(g)
+
\lambda_T \Delta_T(g),
\label{eq:proxy_score}
\end{equation}

where $\Delta_{Y,j}(g) = |\text{Corr}(\hat{x}_j, y) -
\text{Corr}(\tilde{x}_j, y)|$ and $\Delta_{T,j}(g) =
|\text{Corr}(\hat{x}_j, a) - \text{Corr}(\tilde{x}_j, a)|$ measure
how much imputation distorts biomarker--outcome and
biomarker--treatment correlations, with column averages $\Delta_Y$,
$\Delta_T$ (weights $\lambda_Y$ and $\lambda_T$ reported in Table~\ref{tab:llm_hparams}). Penalizing
$\Delta_Y$ directly targets the correlation structure that
CausalFlow-T's do-calculus depends on; penalizing $\Delta_T$ targets
the propensity-related correlations that drive confounding adjustment.

\textbf{Evolutionary loop.}
Starting from a deterministic seed imputer $g^{(0)}$, at 
each iteration $k$ the LLM receives the current-best 
candidate $g^\star_{k-1}$, its score $s(g^\star_{k-1})$, 
and a compact search history summary, then proposes a variation 
$g^{(k)}$. The update rule is
\begin{equation}
  g^\star_k = \begin{cases} 
    g^{(k)} & \text{if } s(g^{(k)}) < s(g^\star_{k-1}), \\ 
    g^\star_{k-1} & \text{otherwise.} 
  \end{cases}
  \label{eq:update}
\end{equation}
After $K$ iterations, $g^\star = g^\star_K$ 
is applied to $\widetilde{\mathcal{D}}$ to produce 
$\hat{\mathcal{D}}$.  We use a
single-parent evolutionary scheme, maintaining a single current-best imputer; implementation details are reported in
Appendix~\ref{app:llm:algo}.

\section{Experiments}
\label{sec:exp}
We evaluate the pipeline in two steps: (i) CausalFlow-T on complete data with known counterfactuals, and (ii) four imputation strategies with CausalFlow-T fixed under 30\%, 50\%, and 80\% MNAR missingness, including three LLM-based variants. Three findings emerge: (1) CausalFlow-T is the only model satisfying all reliability criteria, revealing two distinct failure modes (Section~\ref{sec:exp:causalflowt}); (2) LLM-driven imputation achieves the best pooled performance across biomarker and causal metrics, with GPT-5.4 strongest overall and open-source LLMs demonstrating robustness and reconstruction–causal tradeoffs (Section~\ref{sec:exp:imputation}); (3) the full pipeline recovers a robust per-protocol treatment effects from incomplete real-world EHR data (Section~\ref{sec:realworld}). Experiments were run on a single HPC node (Appendix~\ref{app:compute}).

\subsection{Experimental setup and evaluation protocol}
\label{sec:exp:setup}

\textbf{Datasets.} We evaluate CausalFlow-T on four synthetic datasets
with ground-truth counterfactuals of increasing structural complexity: \textbf{(1)}~Simple 3-Node
($N$=10k, $T$=5; linear, no confounding; positive control);
\textbf{(2)}~LDL Toy ($N$=10k, $T$=5; heterogeneous saturating effect, autoregressive
AR(1) dynamics, selection-on-gain confounding; primary stress-test for
subgroup calibration); \textbf{(3)}~Cox Survival ($N$=30k, $T$=11;
time-dependent hazard, bimodal covariate heterogeneity); and
\textbf{(4)}~CVD Risk Toy ($N$=50k, $T$=10; 17-node DAG, fully mediated
treatment effect via systolic blood pressure (SBP) lags, HR~$\approx$~0.82; hardest benchmark).
For imputation, we use a semi-synthetic dataset built on a real-world EHR backbone~\citep{chmiel2011fire} with ten synthetic
longitudinal biomarkers, a known benchmark ATE ($-3.484$), and MNAR
missingness introduced at 30\%, 50\%, and 80\%. All datasets are described in Appendix~\ref{app:datasets}.

\textbf{Causal inference baselines.} We compare against CVAE, GNN-CVAE,
and TARNet, all sharing the same LSTM temporal encoder; discriminative
sequence models (Causal Transformer, R-MSN, CRN) are excluded by
construction as they fail the three jointly necessary criteria for
distributional counterfactual evaluation (Appendix~\ref{app:baseline_criteria}).

\textbf{Imputation baselines.}
Six imputation strategies are evaluated with CausalFlow-T held fixed: LOCF (carry-forward heuristic), MissForest (non-parametric random-forest), CausalCFM (a DAG-aware flow matching baseline, Appendix ~\ref{app:CFM}), and LLM-driven evolutionary imputation with three backends (GPT-5.4, Qwen3.5-Plus, and GPT-OSS-120b).

\textbf{Evaluation protocols.} CausalFlow-T is validated on four
metrics exposing failure modes invisible to predictive error: subgroup
calibration $|$Bias$_{Q1}|$/MAE$_{Q1}$, where MAE is the mean
absolute error (ratio~$\approx$~1 signals
systematic confounding failure), arm reconstruction error $\text{Err}_a$,
tail variance ratio VR$_{Q4}$ (collapse~$<$~1, overdispersion~$>$~1),
and hazard ratio recovery $|$HR$_{\text{true}} - $HR$_{\text{pred}}|$.
Imputation is evaluated on two complementary layers: biomarker quality
(MAE, root mean squared
error RMSE, autocorrelation error $|\Delta$AC$|$, consecutive-step error
CSE, and correlation inflation $|$inf$_Y|$) and downstream causal quality
(Q1~MAE, $\bar{\text{Err}}_a$, and $|\widehat{\text{ATE}} -
\text{ATE}^*|$), with CausalFlow-T held fixed so differences reflect
imputation alone (formal definitions in Appendix~\ref{app:eval}).

\subsection{CausalFlow-T: Two non-overlapping failure modes}
\label{sec:exp:causalflowt}

Table~\ref{tab:ranks} summarizes per-metric ranks and binary 
reliability criteria across all four synthetic datasets; full 
numerical results per dataset are in 
Appendix~\ref{app:full_results} 
(Tables~\ref{tab:calibration}--\ref{tab:structural}). All models perform comparably on the
positive control (Simple~3-Node), confirming the DAG constraint does
not degrade performance when confounding is absent; two non-overlapping
failure modes emerge on harder benchmarks.
\begin{table}[h]
\centering
\footnotesize
\setlength{\tabcolsep}{2pt}
\caption{Per-metric ranks (1\,=\,best, 5\,=\,worst) across 
four synthetic datasets of increasing structural complexity, 
and binary reliability criteria evaluated across all datasets 
simultaneously. Bias$_{\mathrm{Q1}}$: ratio
$|\mathrm{Bias}_{\mathrm{Q1}}|/\mathrm{MAE}_{\mathrm{Q1}}$.
VR: $|\mathrm{VR}_{\mathrm{Q4}}-1|$.
Arm: mean arm error. HR: $|\mathrm{HR}_{\mathrm{true}}-\mathrm{HR}_{\mathrm{pred}}|$.
Reliability criteria (\cmark/\xmark/--): \textit{Bias}\,${<}0.5$;
\textit{Tail}\,=\,$|\mathrm{VR}_{\mathrm{Q4}}-1|{<}0.5$ on all benchmarks;
\textit{HR}\,=\,correct direction;
\textit{Arm}\,=\,best on ${\geq}2$ benchmarks;
\textit{Stable}\,=\,no explosion or inversion~\citep{austin2015moving}.}
\label{tab:ranks}
\begin{tabular}{l|cc|cccc|ccc|ccc|c|ccccc}
\toprule
 & \multicolumn{2}{c|}{\textbf{Simple}}
 & \multicolumn{4}{c|}{\textbf{LDL}}
 & \multicolumn{3}{c|}{\textbf{Cox}}
 & \multicolumn{3}{c|}{\textbf{CVD Risk}}
 & \multirow{2}{*}{\shortstack{\textbf{Mean}\\\textbf{rank}}}
 & \multicolumn{5}{c}{\textbf{Reliability}} \\
\cmidrule(lr){2-3}\cmidrule(lr){4-7}\cmidrule(lr){8-10}
\cmidrule(lr){11-13}\cmidrule(lr){15-19}
\textbf{Model} & MAE & VR & MAE & Bias & VR & Arm
               & MAE & Arm & HR
               & MAE & VR & HR &
               & Bias & Tail & HR & Arm & Stable \\
\midrule
CausalFlow-T
  & 3 & \textbf{1} & 2 & 2 & \textbf{1} & 3
  & 3 & \textbf{1} & 2
  & \textbf{1} & 2 & \textbf{1} & \textbf{1.83}
  & \cmark & \cmark & \cmark & \cmark & \cmark \\
NF (no DAG)
  & 2 & 2 & 3 & \textbf{1} & 2 & 4
  & 2 & 2 & 4
  & 2 & 5 & 2 & 2.58
  & \cmark & \xmark & \cmark & -- & \xmark \\
GNN-CVAE
  & \textbf{1} & 3 & 5 & 4 & 3 & 5
  & 4 & 5 & 3
  & 3 & 3 & 3 & 3.58
  & \xmark & \xmark & \xmark & \xmark & \xmark \\
CVAE
  & 3 & 5 & 4 & 3 & 4 & 2
  & \textbf{1} & 4 & \textbf{1}
  & 3 & 4 & 4 & 3.17
  & \xmark & \xmark & \xmark & \xmark & \xmark \\
TARNet
  & 4 & 4 & \textbf{1} & 5 & 5 & \textbf{1}
  & 5 & 3 & 5
  & 5 & \textbf{1} & 5 & 3.67
  & \xmark & \cmark & \xmark & -- & \xmark \\
\bottomrule
\end{tabular}
\end{table}

\textbf{Failure mode 1: Confounding separation requires causal structure.} On LDL Toy, GNN-CVAE, CVAE, and TARNet all reach $|\mathrm{Bias}/\mathrm{MAE}|_{\mathrm{Q1}}\approx 1.0$ (every low-effect error is systematic), while NF\,(no~DAG) reduces this to $0.195$ via exact inference alone and CausalFlow-T to $0.270$, both clearing the $0.5$ threshold; CausalFlow-T further achieves $\mathrm{VR}_{\mathrm{Q4}}=1.049\pm 0.033$ (rank~1), the only model preserving tail variance without collapse or inflation. TARNet's lowest absolute MAE ($0.392\pm 0.081$) masks a bias ratio of $\approx 1.0$ across all quartiles and $\mathrm{VR}_{\mathrm{Q4}}=0.639$, illustrating why MAE rank alone is an insufficient criterion. 

\textbf{Failure mode 2: Structural propagation requires 
exact inference.} CVD Risk stress-tests DAG constraints because the treatment effect is entirely mediated, so any model that cannot respect causal ordering will collapse or diverge. GNN-CVAE and CVAE encode the graph yet rely on ELBO approximation, and the resulting inference gap causes posterior collapse to $\mathrm{HR}\approx 1.005$ (null effect); NF\,(no~DAG) performs exact inference but without structural constraint, routing the treatment signal through arbitrary pathways and yielding $\mathrm{HR}=0.834 \pm 0.324$ (clinically meaningless); TARNet, lacking both, inverts the effect entirely ($\mathrm{HR}=1.133\pm 0.148$, predicting harm where there is benefit). Only CausalFlow-T, combining DAG factorization with exact normalizing-flow inference, forces the treatment signal through the correct mediation pathways and recovers $\mathrm{HR}=0.786\pm 0.051$ vs.\ true $0.831$, with arm errors an order of magnitude below all competitors. On the other hand, the dequantization of binary survival outcomes (Cox benchmark) introduces a calibration cost that explains
CVAE's MAE advantage on that benchmark; on structurally meaningful
metrics CausalFlow-T leads with best arm-1 error ($0.014 \pm 0.002$)
and closest HR recovery ($0.866 \pm 0.011$ vs.\ true $0.887$);
discrete flow extensions are a natural direction for future work
(Section~\ref{sec:discussion}).

\textbf{Semi-synthetic validation}.
CausalFlow-T and GNN-CVAE are the only models passing the bias threshold (0.316 and 0.351); CausalFlow-T is further the only model with near-perfect variance preservation ($\mathrm{VR}_{\mathrm{Q4}}\!=\!1.006\pm0.009$) while GNN-CVAE collapses ($\mathrm{VR}\!=\!0.616$), confirming that low systematic error alone is insufficient for individual-level reliability. TARNet achieves the best arm reconstruction ($0.081\pm0.012$ / $0.038\pm0.003$) but fails the bias threshold ($0.708$) and collapses variance ($\mathrm{VR}\!=\!0.608$), reinforcing that factual accuracy does not imply counterfactual reliability. Removing the DAG constraint degrades the bias ratio to $0.883$ with the largest ATE deviation ($-3.382\pm0.251$), underscoring that structural supervision is necessary under realistic clinical distributions. Full results can be found in Appendix~\ref{app:fire_oracle}.

\textbf{CausalFlow-T for causal inference.}
\label{sec:exp:selection}
CausalFlow-T uniquely resolves the bias–variance–calibration tradeoff across all four synthetic datasets (Table~\ref{tab:ranks}; mean rank $1.83$), satisfying all five reliability criteria. Similar results are found when stress-testing in the real-world EHR semi-synthetic dataset. Competing models fail along at least one axis (trading bias for variance collapse, capturing trajectories with subgroup error, or achieving low MAE while reversing treatment direction) and none meets more than two criteria. We therefore fix CausalFlow-T in subsequent imputation experiments so that differences in ATE recovery reflect imputation quality alone.

\subsection{Imputation under MNAR missingness}
\label{sec:exp:imputation}

Full numerical results are in Appendix~\ref{app:imp_results}
(Tables~\ref{tab:biomarker_quality}--\ref{tab:imp_causal}) while
Table~\ref{tab:imp_ranks} summarizes per-metric ranks across 
both evaluation layers.

\begin{table}[h]
\centering
\footnotesize
\setlength{\tabcolsep}{2pt}
\caption{Per-metric ranks (1\,=\,best, 6\,=\,worst) across the three
missingness levels. Biomarker block: pointwise MAE, pointwise RMSE,
lag-1 autocorrelation error $|\Delta\mathrm{AC}|$, consecutive-step error
(CSE), and absolute biomarker--outcome correlation inflation
$|\mathrm{inf}_Y|$. Causal block: Q1 MAE, mean arm reconstruction error
$\bar{\mathrm{Err}}_{a}$, and absolute ATE residual
$|\widehat{\mathrm{ATE}}-\mathrm{ATE}^{*}|$ ($\mathrm{ATE}^{*}=-3.484$).}
\label{tab:imp_ranks}
\begin{minipage}{0.79\linewidth}
\centering
\resizebox{\linewidth}{!}{%
\begin{tabular}{ll|ccccc|ccc|c}
\toprule
 & & \multicolumn{5}{c|}{\textbf{Biomarker quality}}
   & \multicolumn{3}{c|}{\textbf{Downstream causal}}
   & \multirow{2}{*}{\shortstack{\textbf{Mean}\\\textbf{rank}}} \\
\cmidrule(lr){3-7}\cmidrule(lr){8-10}
\textbf{Miss.} & \textbf{Method}
   & MAE & RMSE & $|\Delta\mathrm{AC}|$ & CSE & $|\mathrm{inf}_Y|$
   & Q1 MAE & $\bar{\mathrm{Err}}_{a}$ & $|\Delta\mathrm{ATE}|$ & \\
\midrule
\multirow{6}{*}{30\%}
 & LOCF        & 6 & 6 & 6 & 6 & 6 & 5 & 4 & 4 & 5.38 \\
 & MissForest  & 4 & 4 & 2 & 4 & 5 & 6 & 6 & 6 & 4.62 \\
 & CausalCFM   & 5 & 5 & 5 & 5 & 2 & \textbf{1} & \textbf{1} & 2 & 3.25 \\
 & GPT-5.4     & \textbf{1} & \textbf{1} & 4 & \textbf{1} & 4 & 4 & 3 & \textbf{1} & \textbf{2.38} \\
 & Qwen3.5-Plus        & 3 & 3 & \textbf{1} & 3 & \textbf{1} & 3 & 5 & 5 & 3.00 \\
 & GPT-OSS-120b     & 2 & 2 & 3 & 2 & 3 & 2 & 2 & 3 & \textbf{2.38} \\

\midrule
\multirow{6}{*}{50\%}
 & LOCF        & 6 & 6 & 6 & 6 & 6 & 4 & 3 & \textbf{1} & 4.75 \\
 & MissForest  & 3 & 3 & \textbf{1} & 3 & 5 & 6 & 6 & 5 & 4.00 \\
 & CausalCFM   & 5 & 5 & 3 & 5 & \textbf{1} & 5 & 4 & 3 & 3.88 \\
 & GPT-5.4     & \textbf{1} & \textbf{1} & 2 & \textbf{1} & 3 & 3 & 2 & 4 & \textbf{2.12} \\
 & Qwen3.5-Plus        & 4 & 4 & 5 & 4 & 2 & \textbf{1} & \textbf{1} & 2 & 2.88 \\
 & GPT-OSS-120b     & 2 & 2 & 4 & 2 & 4 & 2 & 5 & 6 & 3.38 \\

\midrule
\multirow{6}{*}{80\%}
 & LOCF        & 6 & 6 & 6 & 5 & 6 & 3 & \textbf{1} & 2 & 4.38 \\
 & MissForest  & 3 & 3 & \textbf{1} & 3 & 2 & \textbf{1} & 4 & 4 & 2.62 \\
 & CausalCFM   & 5 & 5 & 2 & 6 & \textbf{1} & 6 & 6 & 6 & 4.62 \\
 & GPT-5.4     & \textbf{1} & \textbf{1} & 5 & \textbf{1} & 5 & 4 & 3 & \textbf{1} & 2.62 \\
 & Qwen3.5-Plus        & 4 & 4 & 4 & 4 & 3 & 5 & 5 & 5 & 4.25 \\
 & GPT-OSS-120b     & 2 & 2 & 3 & 2 & 4 & 2 & 2 & 3 & \textbf{2.50} \\
\bottomrule
\end{tabular}}
\end{minipage}%
\hfill
\begin{minipage}{0.19\linewidth}
\raggedleft
\begin{tabular}{l|c}
\toprule
\textbf{Method} & \shortstack{\textbf{Mean}\\\textbf{rank}} \\
\midrule
LOCF        & 4.83 \\
MissForest  & 3.75 \\
CausalCFM   & 3.92 \\
GPT-5.4     & \textbf{2.38} \\
Qwen3.5-Plus        & 3.38 \\
GPT-OSS-120b     & 2.75 \\
\bottomrule
\end{tabular}
\end{minipage}
\end{table}

\textbf{Biomarker reconstruction.} GPT-5.4 achieves the lowest
pointwise MAE and RMSE at every missingness level with only modest
degradation from 30\% to 80\% (MAE 1.914$\to$2.059), and the narrowest
normalized error distribution across all missing positions (nMAE 0.55,
0.56, 0.64; Appendix~\ref{app:imp_results}). Qwen3.5-Plus best preserves lag-1
autocorrelation at 30\%, while MissForest is closest at 50\% and 80\%;
CausalCFM best preserves biomarker (outcome correlation at 50\% and 80\%
[$|$inf$_Y|$ $\approx$ 0.01]). GPT-OSS-120b tracks GPT-5.4 closely on
pointwise reconstruction at 50\% and 80\%, supporting robustness of the
LLM-driven strategy beyond a single model.

\textbf{Downstream causal recovery and the reconstruction: causal
tradeoff.} Pointwise reconstruction quality and causal correlation
preservation are not aligned objectives, and Table~\ref{tab:imp_ranks} exposes this
tradeoff directly. At 30\%, GPT-5.4 gives the closest ATE recovery
($|\Delta$ATE$|$ = 0.024) while CausalCFM leads on subgroup and arm
reconstruction; at 50\%, Qwen3.5-Plus gives the strongest causal
reconstruction ($\bar{\text{Err}}_a$ = 0.069) with near-best ATE
residual (0.013 vs.\ 0.007 for LOCF); at 80\%, GPT-5.4 again gives the
closest ATE recovery ($|\Delta$ATE$|$ = 0.013). Above 50\%
missingness (the regime of most real EHR
data~\citep{sun2024incorporating}) LLM imputation is the only
strategy preserving causal correlation structure and ATE recovery where
statistical methods progressively degrade.

\textbf{LLM-driven evolutionary imputation.}
\label{sec:exp:impselection}
LLM-driven imputation achieves the strongest pooled performance across 30–80\% MNAR (mean rank: GPT-5.4 2.38, GPT-OSS-120b 2.75, Qwen3.5-Plus 3.38; MissForest 3.75, CausalCFM 3.92, LOCF 4.83). GPT-5.4 is selected as the default; CausalCFM remains a useful comparator for causal-fidelity diagnostics at lower missingness. Appendix Figure~\ref{fig:llm_search_progress} shows GPT-5.4 search progress, and Appendix Tables~\ref{tab:biomarker_quality}--\ref{tab:imp_causal} compare the final imputer with the first executable GPT-5.4 proposal; bootstrap sensitivity is shown in Appendix Table~\ref{tab:bootstrap_llm}).

\section{Real-world application: Causal inference from healthcare data}
\label{sec:realworld}
\textbf{TTE in incomplete EHR data:} Semi-synthetic benchmarks provide controlled settings with known ATEs, enabling direct evaluation of estimator bias against ground-truth counterfactuals. While our pipeline recovers the ATE under MNAR missingness in this setting, real-world EHR data introduces additional challenges: unlike RCTs, it is collected for clinical and administrative purposes rather than to answer specific causal questions; treatment is not randomized, follow-up is irregular, treatment deviations are informative, and missingness is rarely MAR. To bridge this gap, we apply the two-stage pipeline (LLM imputation 
followed by CausalFlow-T) to the FIRE database, the most comprehensive primary care EHR database in Switzerland~\citep{chmiel2011fire} (Appendix~\ref{app:fire_description}). Using the TTE framework~\citep{Hernan2016PracticeAvailable}, we estimate the per-protocol effect among initiators of GLP-1RAs ($n=2{,}392$) versus SGLT-2is ($n=3{,}722$) on body weight loss (kg) at 1-year in adults with T2D (details in Appendix~\ref{app:real_data}). Causal assumptions are encoded in an expert-based DAG and missing measurements are imputed via our validated LLM-based strategy (Appendix~\ref{app:dag}). Due to privacy and regulatory constraints, the dataset used in this study cannot be publicly released.

\begin{figure}[h]
    \centering
    \includegraphics[width=0.6\textwidth]{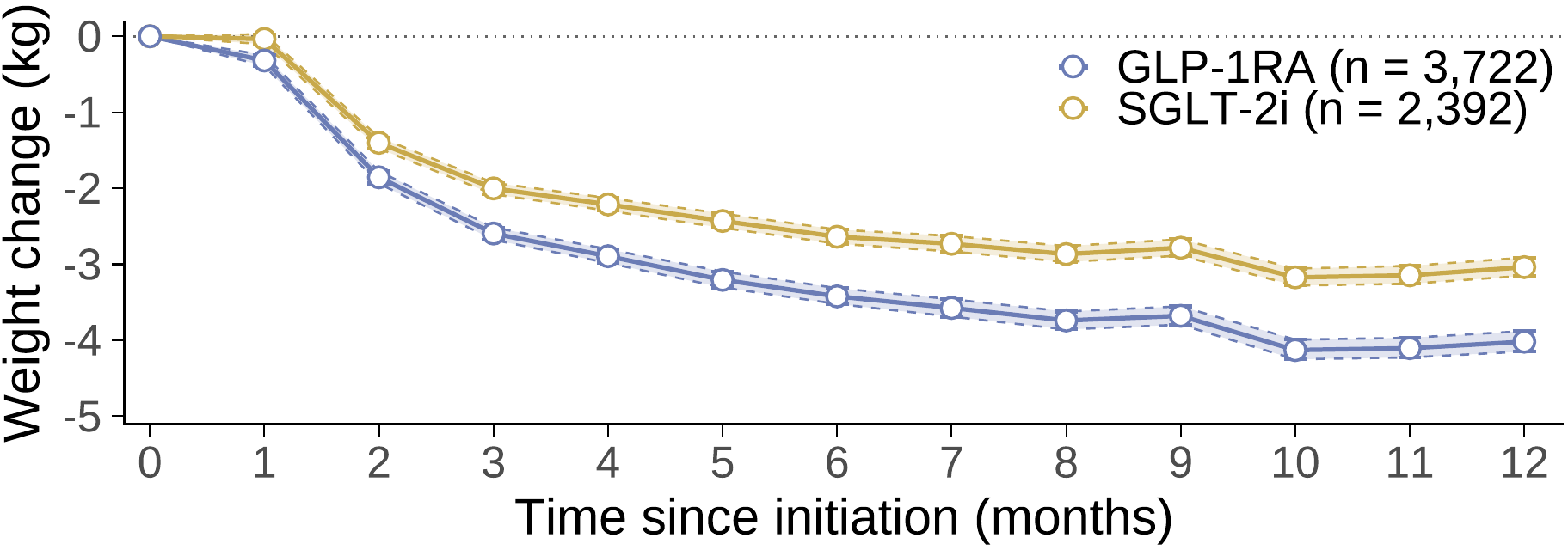}
    \caption{Weight change 
    over 1-year for adults with type 2 diabetes initiating GLP-1 receptor agonists vs
    SGLT-2 inhibitors, estimated from real-world healthcare data using the two-stage pipeline.}
    \label{fig:figure2_neurips2026}
\end{figure}

\textbf{Per-protocol effect estimation and external consistency:} Figure~\ref{fig:figure2_neurips2026} shows both treatment arms exhibit monotonic weight loss from initiation to 1-year, estimated using the proposed pipeline (Appendix~\ref{app:real_data}). At 1-year, the model recovers $\smash{\widehat{\Delta W}}^{\text{GLP-1RA}} -4.02$~kg [$95\%$ CI $-4.15$, $-3.88$] and $\smash{\widehat{\Delta W}}^{\text{SGLT-2i}} -3.04$~kg [$95\%$ CI $-3.15$, $-2.91$], yielding a per-protocol effect of $-0.98$~kg [$95\%$ CI $-1.01$, $-0.96$], favoring GLP-1RAs (bootstrap uncertainty analysis in Appendix Table~\ref{tab:bootstrap_llm}). This estimate is consistent with the SUSTAIN~8 trial (an RCT comparing 
a GLP-1RA [semaglutide] vs (an SGLT-2i [canagliflozin]) in T2D, reporting 1-year body weight difference of $-1.06$~kg (95\%~CI $-1.76$, $-0.36$)~\citep{lingvay2019sustain8}. Our study extends RCT evidence from specific agents in selected trial participants to the full GLP-1RA and SGLT-2i classes used in routine Swiss primary care, where mixed 
agents and $>50\%$ MNAR missingness are expected to attenuate the contrast compared to the SUSTAIN~8 trial estimate~\citep{lingvay2019sustain8}.
While ground-truth counterfactuals are unavailable, consistency with RCT evidence supports the plausibility of the per-protocol effects estimated by the proposed pipeline from incomplete real-world EHR data.

\section{Discussion}
\label{sec:discussion}
This work addresses a key gap in causal inference 
from observational data: jointly handling time-varying 
confounding and high-rate MNAR missingness, two compounding 
challenges that may substantially distort treatment effect 
estimation in observational data. We propose a two-stage pipeline in which CausalFlow-T, a DAG-constrained normalizing flow with exact 
counterfactual inference, is validated on complete data 
before an LLM-driven evolutionary imputer enables deployment under realistic missingness, isolating each component's performance impact.

A central finding is not that CausalFlow-T outperforms competing models, but 
that DAG constraints and exact inference address strictly 
non-overlapping failure modes, neither compensates for 
the absence of the other. Existing methods 
evaluated on simpler benchmarks may be systematically 
overestimating their robustness: models that perform well on factual prediction can still fail on structurally meaningful criteria such as subgroup calibration, tail variance, and mediation-consistent effect propagation. Our five-criterion reliability framework provides a more stringent evaluation template than any single metric and may serve as a useful benchmark for future work in longitudinal causal inference.

The imputation results reveal a complementary insight: pointwise reconstruction and causal correlation preservation are not aligned objectives. LLM-based methods, especially GPT-5.4, dominate on
reconstruction and temporal extrapolation, while CausalCFM performs better on selected causal-fidelity diagnostics, especially biomarker--outcome
correlation preservation. This tradeoff is invisible
to standard imputation benchmarks, motivating joint biomarker--causal
evaluation as a necessary standard for causal pipelines under
missingness. Future methods could explicitly optimize this tradeoff instead of separating reconstruction from causal fidelity.

Applied to $6{,}114$ adults with T2D in 
Swiss primary care, the pipeline recovers a per-protocol effect of 
$-0.98$~kg [$95\%$~CI $-1.01$, $-0.96$] favoring 
GLP-1RAs over SGLT-2is, consistent with RCT evidence~\citep{lingvay2019sustain8}. Although ground-truth counterfactuals are unavailable, this suggests that the proposed pipeline recovers clinically plausible treatment effect estimates from incomplete EHR data, extending causal evidence beyond selected participants and conditions typical for RCTs.

Several limitations suggest directions for future work.
\label{sec:limits}
First, CausalFlow-T uses dequantization for binary outcomes, incurring calibration costs on survival endpoints that explain CVAE’s lower Cox Survival MAE despite poorer hazard-ratio recovery and arm reconstruction (Appendix~\ref{app:theory}); 
discrete or hybrid flow-based architectures for survival outcomes are a 
natural extension. Second, the DAG is assumed fixed and expert-specified. Clinically motivated single-edge removals reveal a bias–instability tradeoff: deleting unique causal pathways primarily increases bias, whereas deleting redundant pathways increases estimation instability, suggesting that cross-seed instability may provide a differentiable signal for data-driven DAG refinement without ground-truth counterfactuals (Appendix ~\ref{app:dag_sensitivity}). 
However, edge-direction errors and latent confounding remain important limitations for future work. Third, although we evaluate three LLM backends, all use the same
evolutionary search protocol, and performance may therefore depend on the search design. Future work should explore alternative evolutionary strategies, including recombination and multi-objective acceptance rules. Finally, like all observational studies, 
our real-world data analysis assumes conditional exchangeability given the 
expert-specified adjustment set, a standard but untestable assumption. Future work could incorporate sensitivity analyses 
for unmeasured confounding. 

\section{Conclusion}
We introduced a two-stage pipeline for treatment effect estimation from 
incomplete longitudinal EHR data, combining CausalFlow-T (a 
DAG-constrained normalizing flow for exact counterfactual inference) with an LLM-driven 
evolutionary imputer for MNAR missingness. 
Controlled ablations show that DAG constraints and exact inference address distinct failure modes in longitudinal causal inference, with CausalFlow-T the only model satisfying all five reliability criteria simultaneously. Joint biomarker–causal evaluation further reveals causal implications of imputation choices that reconstruction-only metrics overlook. Applied to real-world EHR data, the pipeline recovers a per-protocol effect consistent with RCT evidence despite substantial MNAR missingness. 
Our findings demonstrate that jointly addressing causal structure and MNAR missingness is both necessary and feasible, supporting ML-based treatment effect estimation from observational data as a practical complement to RCTs and extending causal inference to real-world populations and care settings where trials are infeasible or insufficient and real-world evidence is needed to inform clinical and policy decision-making.

\newpage
\bibliographystyle{plainnat}
\bibliography{references}


\newpage
\appendix
 
\section{Dataset Descriptions}
\label{app:datasets}

All four benchmarks are fully synthetic, so both ground-truth outcomes $Y(0)$ and $Y(1)$ are available for every test patient,
enabling exact evaluation of counterfactual predictions.
Train/test splits of 80/20 are used throughout.
 
\subsection{Simple 3-Node}
\label{app:datasets:simple}

The dataset contains $N$ patients observed over $T$ timesteps (default
$N{=}10{,}000$, $T{=}5$).  Three variables are simulated:

\begin{itemize}
  \item \textbf{Confounder} $X_1^{(t)} \sim \mathcal{N}(t/4,\,1)$:
        a single continuous covariate whose mean drifts linearly with
        time, inducing non-stationarity.
  \item \textbf{Treatment} $T \in \{0,1\}$: a binary treatment assigned once at baseline and kept constant.  The assignment
        probability is a nonlinear function of $X_1$:
        \[
          P(T=1 \mid X_1) =
          \begin{cases}
            0.2 & \text{if } \bar{X}_1 > 2.5, \\
            0.8 & \text{otherwise,}
          \end{cases}
        \]
        where $\bar{X}_1 = \mathbb{E}_t[X_1^{(t)}]$ and the
        intermediate variable $X_1^2 - \sin(X_1) + \varepsilon$
        ($\varepsilon \sim \mathcal{N}(0, 0.25)$) mediates the
        propensity.  This creates nonlinear confounding with
        a treatment probability that flips between 20\% and 80\%.
  \item \textbf{Outcome} $Y^{(t)}$: a continuous outcome generated as
        \[
          Y^{(t)} =
          \begin{cases}
            3X_1^{(t)} + 0.25\,(t/T) + \varepsilon_0 & T = 0,\\
            3X_1^{(t)} - 0.50\,(t/T) + \varepsilon_1 & T = 1,
          \end{cases}
        \]
        where $\varepsilon_0, \varepsilon_1 \sim \mathcal{N}(0,1)$.
        Treatment reduces the time trend by $0.75\,(t/T)$ relative
        to control, inducing a modest, time-growing ITE.
\end{itemize}

\paragraph{Causal graph.}
The DAG is $X_1 \to T \to Y$, $X_1 \to Y$, with self-loops at each
timestep, encoded as a $3{\times}3$ adjacency matrix.

\paragraph{Purpose.}
A positive control with a low confounded structure (linear dependencies mostly). Because the causal ordering matches the variable index order and confounding is low, all five models are expected to perform comparably. The dataset verifies that the framework performs well even in the simplest case.
 
\subsection{LDL Toy}
Generated dataset with $N{=}10{,}000$ patients, $T{=}5$ timesteps.  The outcome $Y$ represents a continuous biomarker (analogous to LDL cholesterol) subject to a heterogeneous drug response, log-saturation pharmacokinetics, autoregressive dynamics, and selection-on-gain confounding.

Specifically, the response magnitude at baseline is
$\delta_i \sim \mathrm{Uniform}(0.20, 0.60)$, giving each patient an
individual treatment-effect scale (20\%--60\% reduction).  The
dose-response saturates over time according to
\[
  \mathrm{sat}(t)
  = \frac{\log(1 + \lambda t)}{\log(1 + \lambda T_{\max})},
  \quad \lambda = 0.4,
\]
so early timesteps carry most of the treatment signal.  
Temporal dependence is introduced via a first-order AR(1) process with coefficient $\rho = 0.75$, such that each observation exhibits strong dependence on its immediate predecessor.

Confounding is selection on gain: patients who respond more
strongly ($\delta_i$ above median) are preferentially assigned to
treatment, so the treated group has a higher true effect than the
control group. 

\paragraph{Purpose.}
Tests calibration of heterogeneous treatment effects.  The combination of log-saturation (non-constant effect over time), AR(1) dynamics (temporal correlation), and selection-on-gain confounding makes this the primary benchmark for the structural-failure and variance-collapse diagnostics.

\subsection{Cox Survival Toy}
The dataset contains $N{=}30{,}000$ patients, $T{=}11$ timesteps (including $t{=}0$), $p{=}5$ covariates.

\begin{enumerate}
  \item \textbf{Covariates.}  Five covariates are drawn from bimodal
        Gaussian mixtures:
        \[
          X_{ij} \mid Z_{ij} =
          \begin{cases}
            \mathcal{N}(\mu_1^{(j)},\, 0.01) & Z_{ij}=1,\\
            \mathcal{N}(\mu_2^{(j)},\, 0.04) & Z_{ij}=0,
          \end{cases}
        \]
        with $Z_{ij} \sim \mathrm{Bernoulli}(0.5)$ and
        $(\mu_1^{(1)},\ldots)=(3,1,2,5,0)$,
        $(\mu_2^{(1)},\ldots)=(0,2,4,5,5)$.

  \item \textbf{Hazard function.}  Per-patient log hazard ratios are
        \[
          \log \mathrm{HR}_i = \boldsymbol{\beta}^\top X_i,\quad
          \boldsymbol{\beta} = (0.2,\,-0.4,\,-0.3,\,0.3,\,-0.5),
        \]
        giving $\mathrm{HR}_i = \exp(\boldsymbol{\beta}^\top X_i)$.
        The baseline hazard is $h_0 = 0.5$.

  \item \textbf{Treatment assignment (confounding).}
        Patients with $\mathrm{HR}_i \geq 0.5$ (i.e., higher risk)
        are treated with probability 0.70; the remainder with
        probability 0.30.

  \item \textbf{Treatment effect.}  Treatment reduces the individual
        hazard ratio by 30\%:
        $\mathrm{HR}_i^{\text{treated}} = 0.70\,\mathrm{HR}_i$.

  \item \textbf{Event times.}  Survival times are drawn from an
        exponential distribution,
        $\mathcal{E}(1/[h_0\,\mathrm{HR}_i])$,
        and converted to a binary event indicator at each of the 11
        timesteps.  Censoring is applied only at the end of the study (all weight is on the final timestep), so within-study censoring is
        minimal.

  \item \textbf{Counterfactuals.}  The smooth individual cumulative
        incidence function
        $F_i(t) = 1 - \exp(-h_0\,\mathrm{HR}_i\,t)$
        is computed for both the always-treated and never-treated
        hazards and stored as ground-truth potential outcomes for
        evaluation.
\end{enumerate}

\paragraph{Purpose.}
Tests survival-model calibration under continuous confounding.
The bimodal covariate structure creates two patient clusters. The correct arm reconstruction requires the model to disentangle hazard ratio heterogeneity from the selection-on-risk confounding.

\subsection{CVD Risk Toy}
\label{app:datasets:cvd}

The CVD Risk Toy is the most complex benchmark, designed to emulate a
real-world antihypertensive target-trial setting. It is semi-synthetic:
the causal structure, covariate relationships, treatment dynamics, and
outcome model are grounded in established cardiovascular epidemiology
and RCT evidence, while outcomes are re-generated under a fully
specified data-generating process to provide a known ground truth (including exact counterfactual trajectories and a precise hazard ratio
($\mathrm{HR}\approx 0.82$)). The dataset comprises $N{=}50{,}000$
patients over $T{=}10$ observation years, with 15 time-varying
covariates and a 17-node DAG encoding the full mediation structure.

\begin{enumerate}
  \item \textbf{Baseline demographics.}
        Age $\sim \mathcal{N}(55, 100)$; BMI from age with Gaussian
        noise; total cholesterol (TC) from a log-normal with age/BMI
        dependence; high-density lipoprotein (HDL) inversely related to age and BMI; smoking
        status $\sim \mathrm{Bernoulli}(0.6)$; diabetes probability
        proportional to BMI; race and sex drawn from fixed population
        proportions.
  \item \textbf{Time-varying dynamics.}
        At each year $t$, age increments by 1; BMI, TC, and HDL evolve
        with Gaussian noise; hypertension ($\mathrm{htn}$) transitions
        to 1 irreversibly with a probability driven by age, TC, and
        smoking; diabetes is absorbing once acquired.
        Systolic blood pressure (SBP) at time $t$ is:
        \[
          \mathrm{SBP}^{(t)} = 70\,\frac{\mathrm{SBP}^{(t-1)}}
          {\overline{\mathrm{SBP}}_0}
          + 0.5\,\mathrm{age}^{(t)} + 0.15\,\mathrm{TC}^{(t)}
          + 10\,\mathrm{smoker} + \varepsilon,\;
          \varepsilon\sim\mathcal{N}(0,400),
        \]
        clipped to $[80, 200]$.
  \item \textbf{Treatment assignment (absorbing).}
        At $t{=}0$, patients with hypertension are treated with
        probability 0.70; non-hypertensive patients with probability
        0.30. Treatment is absorbing: once assigned it persists.
        At subsequent time steps, untreated hypertensive patients
        initiate treatment with probability 0.01 per year.
  \item \textbf{Mediated treatment effect.}
        The antihypertensive reduces SBP via lagged mediation:
        \[
          \mathrm{SBP}_{\mathrm{final}}^{(t)} = \mathrm{SBP}^{(t)}
          \bigl(1 - [0.20\,T^{(t-1)} - 0.05\,T^{(t-2)}
                      - 0.02\,T^{(t-3)}]\bigr),
        \]
        with lags 1-3. This is the only pathway through which
        treatment affects the outcome; the complete causal chain is
        $T \to \mathrm{SBP}_{\mathrm{final}} \to Y$.
  \item \textbf{CVD outcome.}
        Binary CVD event at each timestep from a logistic model
        with 12 risk-factor coefficients:
        \begin{align*}
          \log\frac{p}{1-p} &= -10
            + 0.005\,\mathrm{age}
            + 0.15\,\mathrm{sex}
            + 0.03\,\mathrm{BMI}
            + 0.015\,\mathrm{SBP}_{\mathrm{final}} \\
          &\quad - 0.01\,\mathrm{HDL}
            + 0.01\,\mathrm{TC}
            + 0.25\,\mathrm{smoker}
            + 0.30\,\mathrm{diabetes} \\
          &\quad + 0.20\,\mathrm{fam\_hx}
            + 0.10\,\mathbb{1}[\mathrm{race=Black}]
            - 0.05\,\mathbb{1}[\mathrm{race=Asian}].
        \end{align*}
        The outcome is absorbing: rows after the first event are
        excluded. Both always-treated and never-treated probability
        trajectories are stored as counterfactual targets.
\end{enumerate}

\paragraph{17-node DAG.}
The causal graph encodes the full mediation structure:
$T \to \mathrm{SBP}_{\mathrm{final}} \to Y$, plus direct paths from
all 15 covariates to $Y$ and temporal self-edges.

\paragraph{Purpose.}
Strong confounding (hypertensive patients preferentially treated), a
fully mediated treatment pathway (SBP lags 1-3), 15 correlated
time-varying covariates, and an absorbing binary outcome combine to
make this the hardest benchmark. Correctly recovering the true
$\mathrm{HR}\approx 0.82$ requires disentangling the mediation chain
from the confounded observational distribution (precisely the
setting where the expert DAG prior is expected to matter most).

\subsection{Summary and main purpose}
The four benchmarks vary deliberately in complexity.

\emph{Simple 3-Node} is a positive control (minimal confounding,
three variables, known closed-form ITE) and is used to verify that all the models perform well in a simple, unrealistic case.

\emph{Cox Survival Toy} introduces a survival endpoint with
bimodal covariate heterogeneity and selection-on-risk confounding,
testing whether models correctly recover per-arm cumulative incidence functions under censoring.

\emph{LDL Toy} and \emph{CVD Risk Toy} are the two hardest
benchmarks and carry the most diagnostic weight in our evaluation.
LDL Toy combines log-saturation pharmacokinetics, AR(1) temporal
dynamics, and selection-on-gain confounding, making it the primary
stress-test for subgroup calibration: a model that merely minimizes
factual mean squared error (MSE) will exhibit structural bias (bias/MAE\,$\approx$\,1) and variance collapse (VR$_{Q4}$\,$<$\,1) under this design.

On the other hand, CVD Risk Toy is the most demanding benchmark overall: 15 time-varying covariates, an absorbing binary outcome, a purely mediated treatment effect through SBP (lags 1--3), and a 17-node expert DAG create a setting where correctly recovering the interventional distribution $p(Y \mid do(T))$ requires explicit encoding of the causal graph.
It is the only benchmark where direction inversion
($\widehat{\mathrm{HR}} < 0$ or $> 1$) is observed for models lacking structural constraints, making it the definitive test of causal identifiability.

Together, this benchmark aims to cover the three main failure modes of longitudinal causal models (confounding, variance miscalibration, and mediation) while providing exact ground-truth potential outcomes for all evaluation metrics.

\subsection{Real-World EHR Database Description and Use}
\label{app:fire_description}
The Family Medicine Research using Electronic Medical Records (FIRE) database (managed by the Institute of Primary Care at the University of Zurich, at the University Hospital Zurich, Switzerland) is the most comprehensive Swiss 
primary-care database of longitudinal (time-stamped) EHR from general practitioners across 
Switzerland.~\citep{balaj2025leveraging} The FIRE database contains individual-level information on demographics, consultations, medication prescriptions, 
laboratory values, vital signs, and diagnoses~\citep{balaj2025leveraging, chmiel2011fire}. FIRE is used in two ways throughout this work: 
as a clinical backbone for generating the semi-synthetic 
evaluation dataset (Section~\ref{app:data}), and to create the study population for the real-world application
(Section~\ref{app:real_data}).
\label{sec:semi_synthetic}

\subsubsection{Semi-Synthetic MNAR Dataset}
\label{app:data}
The semi-synthetic evalution dataset is built on a patient-month panel structure from FIRE, keeping all observed real-world information on
demographics, comorbidities, comedication, and calendar time.

\paragraph{Synthetic covariates:} 
Ten continuous longitudinal variables are generated recursively on top of this backbone, 
indexed by patient and by month. The variables share latent factors and 
cross-variable dependencies, but in order to be realistic, the variables are designed to be non-smooth 
across time: they rely on near-memoryless shared latent drivers, external 
shocks, and noise. 

The generation proceeds as follows:
\begin{enumerate}
    \item A patient-month panel is constructed from selected demographic and clinical covariates.
    \item Within-patient time indices are created.
    \item Three patient-level latent traits are drawn once per patient.
    \item Two weakly autocorrelated shared state variables drive autocorrelation across series.
    \item External shocks introduce abrupt level changes.
    \item Ten variable-specific noise terms complete the stochastic structure.
    \item For each patient, a recursive simulation initializes each variable from covariates, latent traits, and noise, then iterates forward introducing autocorrelation, cross-variable dependence, treatment indicators, latent states, and shocks.
\end{enumerate}

\paragraph{Synthetic outcome:}
The continuous outcome is a nonlinear function of the 
synthetic covariates, clinical covariates, treatment 
assignment, time gap, and lagged terms, with added 
individual noise and occasional outliers.

\paragraph{Ground-truth counterfactuals and benchmark average treatment effect (ATE):}
Potential outcomes under both treatment arms are computed directly from the structural outcome function, providing exact individual-level counterfactuals and a known benchmark ATE.

\paragraph{MNAR missingness mechanism:}
Missingness is introduced into the ten synthetic 
covariates at 30\%, 50\%, and 80\% following the MNAR 
mechanism specified in Equation~\ref{eq:mnar}. This 
mirrors the clinical pattern where measurements are 
triggered by patient condition rather than scheduled, meaning the probability of observing a value 
depends on the underlying patient state. This controlled setup allows us to directly evaluate imputation quality and the subsequent robustness of causal estimators against a known ground truth. 

\begin{figure}[h]
    \centering
    \includegraphics[width=1\linewidth]{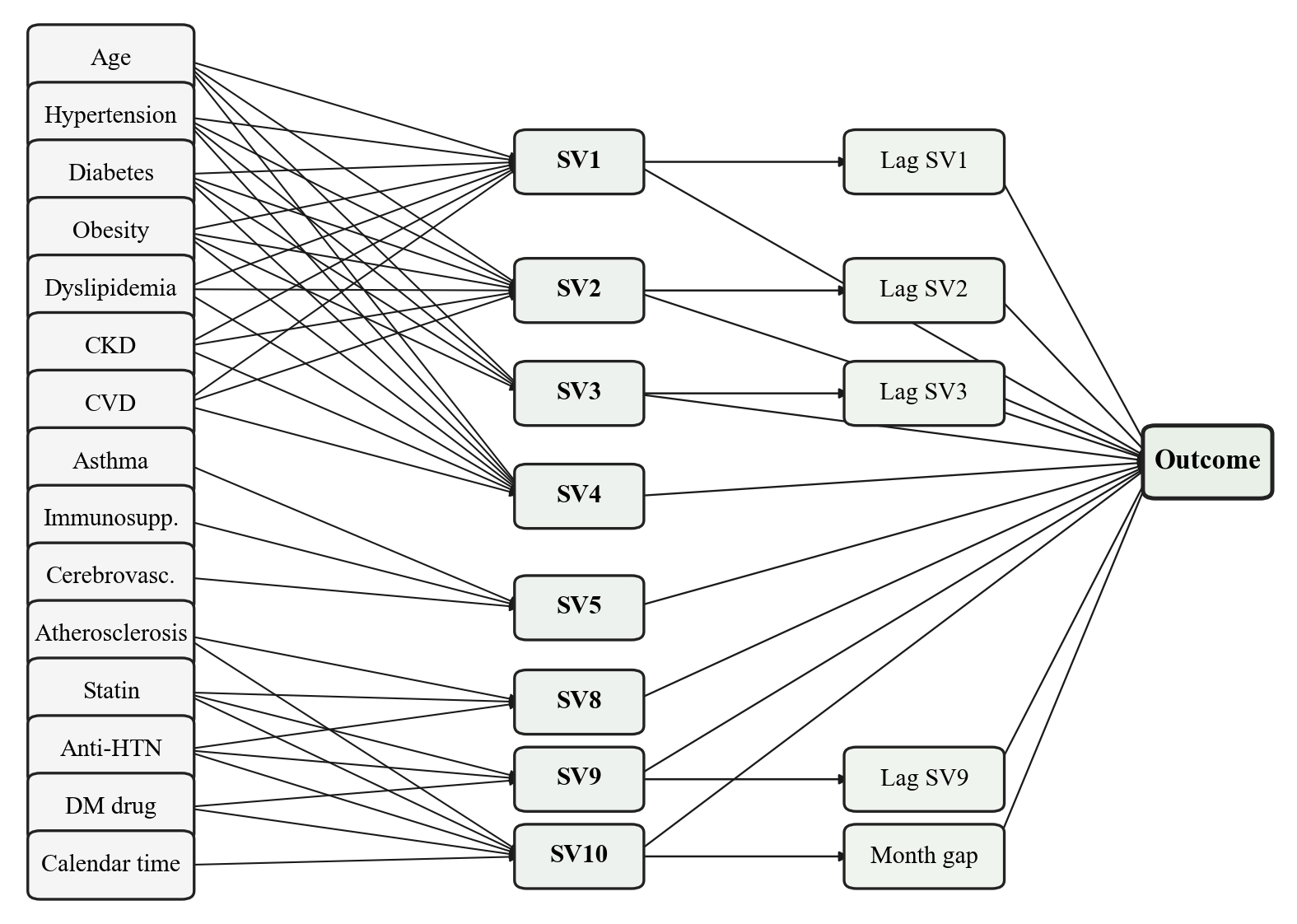}
    \caption{Directed-Acyclic-Graph of the generation of the synthetic variables and the outcome.}
    \label{fig:DAG_SV}
\end{figure}

\begin{figure}[h]
    \centering
    \includegraphics[width=1\linewidth]{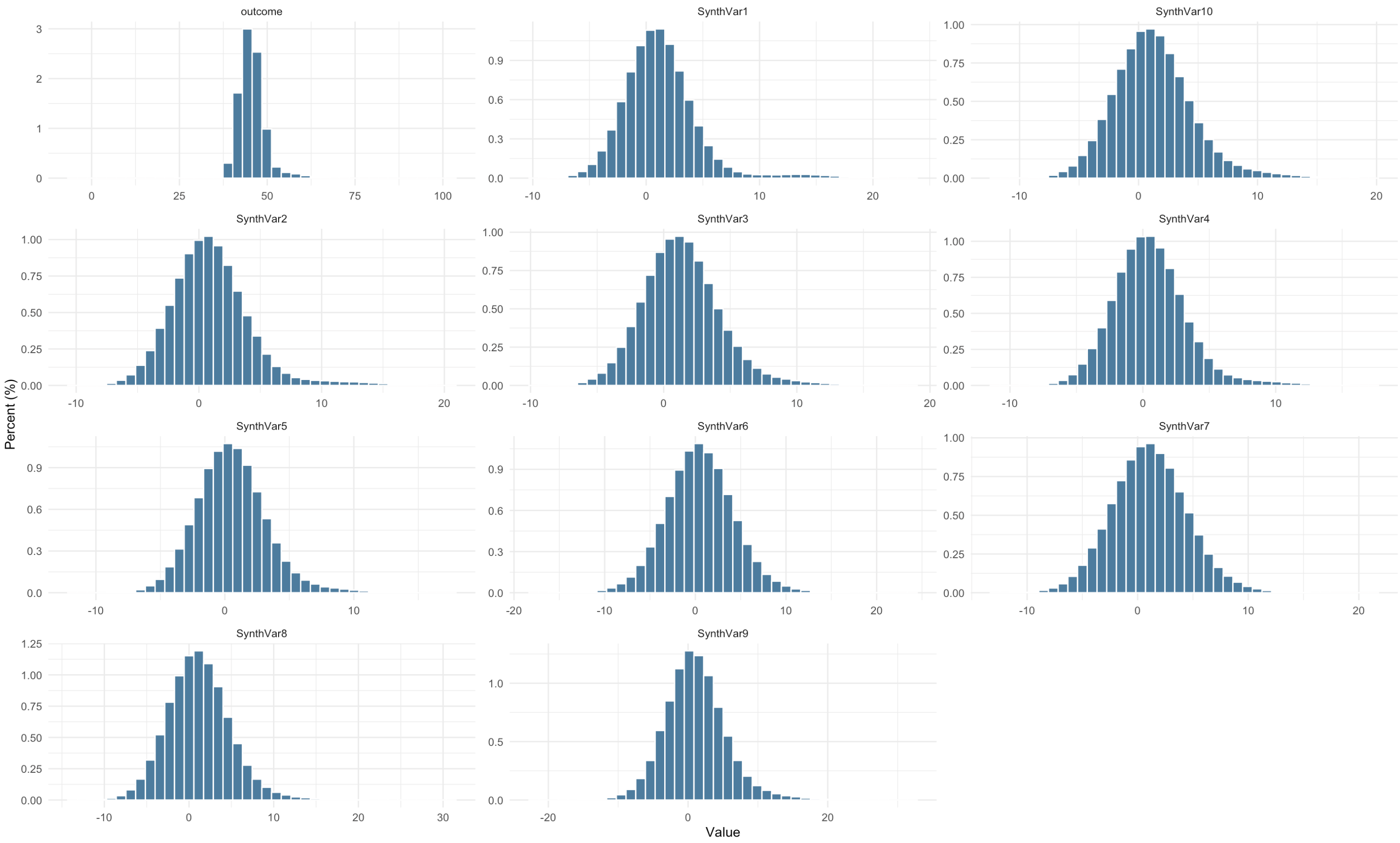}
    \caption{Distribution of the synthetic variables and outcome.}
    \label{fig:dist_SV}
\end{figure}

\paragraph{Directed acyclic graph (DAG) and variable distributions:}
Figure~\ref{fig:DAG_SV} illustrates the DAG of the 
generative process. We also show the resulting distribution of the synthetic variables in Figure~\ref{fig:dist_SV}. Lagged covariates are included for 
realism: \texttt{lag(Synthetic Variable)} at time $t$ 
is derived from the same variable at time $t-1$, so 
the causal direction is past value $\to$ lagged feature 
at the current step. Although the naming may seem counterintuitive, the causal direction is straightforward: the past value of a variable drives its own lagged feature at the current time step. Figure~\ref{fig:dist_SV} shows 
the resulting variable distributions. 

\subsubsection{Target trial emulation using Real-World EHR data}
\label{app:real_data}

\textbf{Study design and patient cohort construction:} 
Using the FIRE database, we design an active-comparator, new-user cohort study (emulating a target trial; see Table~\ref{tab:tte_spec}) to estimate the effect of initiating a GLP-1 receptor agonist (GLP-1RA) or an SGLT-2 inhibitor (SGLT-2i) on body weight change from treatment initiation over 1-year. Cohort entry (time-zero) is defined as the date of the first-ever prescription for a GLP-1RA or an SGLT-2i as first second-line glucose-lowering therapy between 01-Jan-2015 and 07-Dec-2024. To be eligible, individuals are aged 18-years or older at cohort entry and have a diagnosis of type 2 diabetes (T2D) before or at cohort entry (see detailed eligibility criteria in Table~\ref{tab:tte_spec}). 

\textbf{Treatment strategies and follow-up:} We use a per-protocol analysis in which eligible 
individuals are followed from cohort entry until the 
earliest of: deviation from the assigned treatment 
strategy (defined as a gap between consecutive 
prescriptions exceeding 365-days, with a 30-day grace 
period); initiation of the comparator drug class; 
death; or end of the pre-specified follow-up period 
(390 days after cohort entry). 

\textbf{Outcome:} The outcome is the absolute change in bodyweight (kg) from baseline over 1-year. The outcome is extracted in 30-day intervals throughout follow-up. 

\textbf{Covariates:} Time-fixed baseline covariates measured at cohort 
entry include demographic characteristics (age, sex 
assigned at birth). Time-varying covariates updated from baseline every 30-day 
interval throughout follow-up include current comedications 
(antihypertensives, statins), comorbidities 
(cardiovascular disease, chronic kidney disease, 
hypertension, dyslipidemia), and laboratory 
measurements (Hemoglobin A\textsubscript{1c}, eGFR, LDL 
cholesterol, systolic blood pressure). 

\begin{figure}[h]
    \centering
    \includegraphics[width=0.9\linewidth]{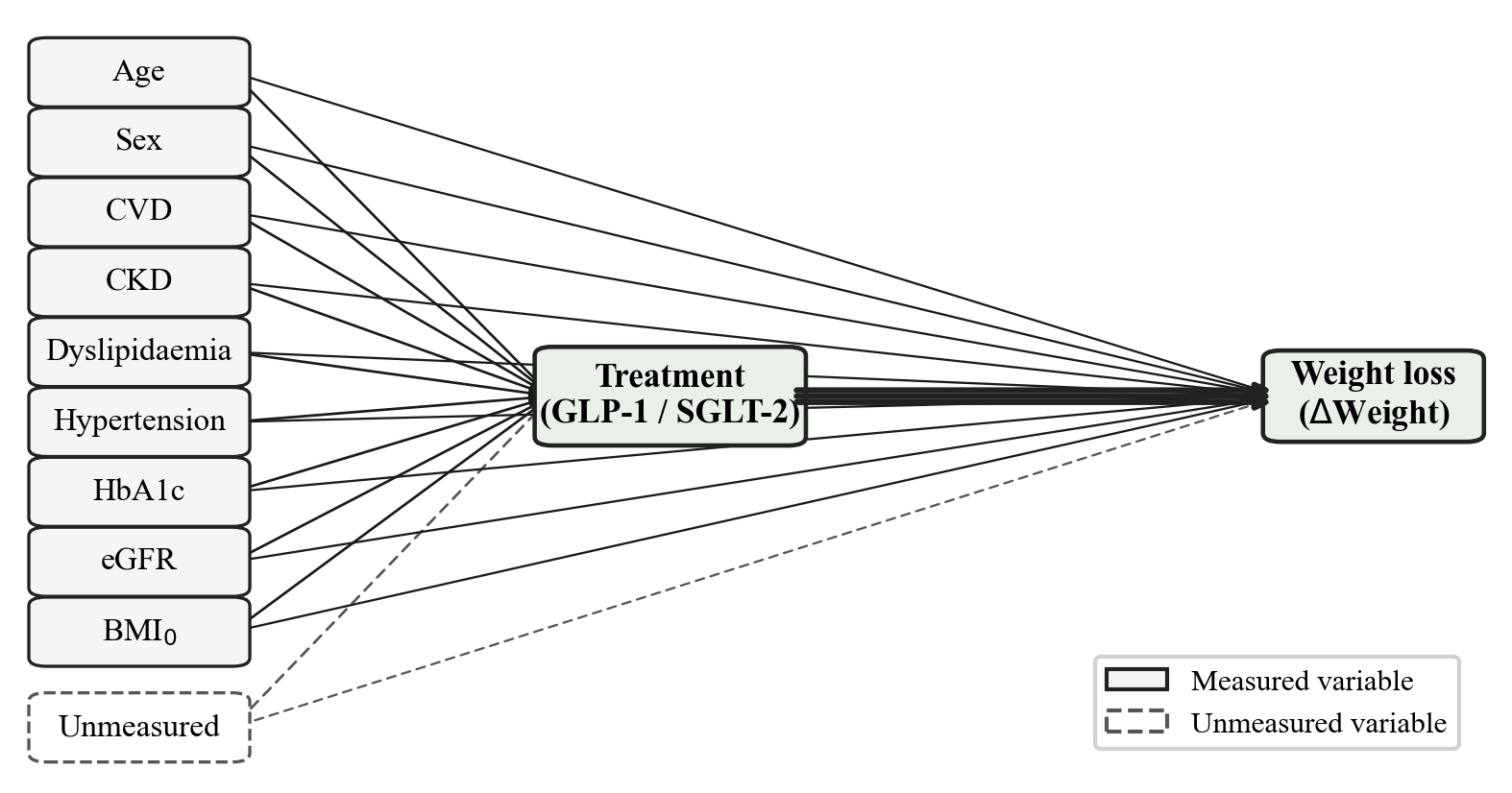}
    \caption{Expert-specified directed acyclic graph for the active-comparator, new-user cohort study assessing the effect of GLP-1 
    receptor agonist versus SGLT-2 inhibitor initiation on weight loss. The DAG is a simplified version with no differentiation between baseline and time varying-confounders.}
    \label{fig:DAG_glp1}
\end{figure}

\textbf{Confounders and DAG:}
\label{app:dag}
Using the covariates defined above, causal assumptions are encoded in an expert-specified 
DAG developed with pharmacoepidemiology 
experts (Figure~\ref{fig:DAG_glp1}), following 
standard practice in target trial 
emulation~\citep{hernan2022target}. The DAG identifies 
the minimal sufficient adjustment set required for 
conditional exchangeability between treatment arms.

\textbf{Real-world application of the proposed ML pipeline:}
The final cohort comprised $6{,}114$ adults with T2D 
initiating a GLP-1RA ($n = 2{,}392$) 
or an SGLT-2i ($n = 3{,}722$) as their 
first second-line glucose-lowering therapy. Following our proposed pipeline (section~\ref{sec:overview}), missing 
covariates or outcome measurements are imputed using the 
validated LLM-based strategy across follow-up. CausalFlow-T was then fitted on the 
completed longitudinal panel.

\begin{table}[htbp]
\centering
\footnotesize
\renewcommand{\arraystretch}{1.45}
\setlength{\tabcolsep}{5pt}
\caption{Specification and emulation of the target trial emulation protocol aiming to estimate the per-protocol effect of GLP-1 receptor agonist (GLP-1RA) versus SGLT-2 inhibitor (SGLT-2i) initiation on body weight at one year in adults with type 2 diabetes (T2D).}
\label{tab:tte_spec}
\begin{tabularx}{\linewidth}{
  >{\bfseries\raggedright\arraybackslash}p{2.5cm}
  >{\raggedright\arraybackslash}X
  >{\raggedright\arraybackslash}X}

\toprule
Protocol component
  & Target trial specification
  & Target trial emulation \\
\midrule

Eligibility criteria
  & Inclusion criteria: Adults ($\geq$18 yrs) initiating a GLP-1RA or an SGLT-2i; T2D diagnosis.

  Exclusion criteria: Prescription for the comparator drug class before treatment initiation; type 1 diabetes diagnosis; last recorded eGFR $\leq$30ml/min/1.73m2.

  & Inclusion criteria: Adults ($\geq$18 yrs) initiating a GLP-1RA or an SGLT-2i between 01-Jan-2015 and 07-Dec-2024 (cohort entry = first prescription date); T2D diagnosis before or at cohort entry; at least one weight measurement recorded within the 365 days before cohort entry.

  Exclusion criteria: Prescription for the comparator drug class before or at cohort entry; type 1 diabetes diagnosis before or at cohort entry; eGFR $\leq$30 before or at cohort entry. Missingness in variables need to assess eligibility are imputed using the LLM-based strategy prior to eligibility
    assessment. \\

    \addlinespace
Treatment strategies
  & (1) Initiate and sustain a GLP-1RA receptor 
    agonist; (2) initiate and sustain an SGLT-2i 
    inhibitor; as add-on to first-line 
    glucose-lowering therapy over 12 months.
  & Same strategies; patients classified by 
    observed prescription at cohort entry. \\

    \addlinespace
Treatment assignment
  & Random assignment at baseline (open-label); 
    participants not blinded to assigned strategy.
  & Non-random; conditional exchangeability 
    emulated by adjusting for the minimal 
    sufficient adjustment set identified from 
    the expert-specified DAG (Figure~\ref{fig:DAG_glp1}). \\

\addlinespace
Outcome
  & Change in body weight (kg) from baseline to 1-year (12 months), measured in 30-day intervals throughout follow-up.
  & Same. Missing outcome measurements are imputed using the LLM-based strategy.\\

\addlinespace
Follow-up
  & From baseline until the earliest of: 
    per-protocol deviation from assigned 
    strategy; initiation of comparator drug 
    class; death; or 390 days after cohort 
    entry.
  & Same. Per-protocol deviation defined as 
    a gap between consecutive prescriptions 
    exceeding 365 days (and adding a 30-day grace period) 
    or initiation of the comparator drug class. \\

\addlinespace
Causal contrast
  & Per-protocol effect: treatment effect for all randomized individuals if all adhered to treatment and did not initiate the an agent of the comparator drug class.
  & Observational analog of the per-protocol 
    effect, estimated under the assumption of 
    conditional exchangeability given the 
    adjustment set identified from the DAG(Figure~\ref{fig:DAG_glp1}). \\

\addlinespace
Statistical analysis
  & ATE estimation via marginal structural 
model with inverse probability of censoring weighting to account for informative 
censoring due to non-adherence.
  & Following our proposed pipeline (section~\ref{sec:overview}), missing 
covariates or outcome measurements are imputed using the 
validated LLM-based strategy across follow-up. CausalFlow-T was then fitted on the 
completed longitudinal panel. \\

\bottomrule
\end{tabularx}
\end{table}

\section{Baseline Architectures for Causal Inference Estimation}
\label{app:baselines}
\subsection{Baseline Selection Criteria}
\label{app:baseline_criteria}
We restrict causal inference baselines to generative distributional 
models satisfying three jointly necessary criteria:

\begin{enumerate}
    \item \textbf{Generative mechanism.} The model must define a 
    joint distribution $p(Y^{(0)}, Y^{(1)} \mid X)$ rather than 
    only conditional means $\mathbb{E}[Y \mid X, A]$, as the latter 
    is insufficient for individual-level counterfactual inference 
    via the AAP procedure~\citep{pearl2009}.
    
    \item \textbf{Invertible abduction.} The model must support 
    exact or approximate recovery of patient-specific exogenous 
    noise $z$ from observations, implementing the twin-network 
    assumption required for individual-level counterfactuals; 
    without this, $\hat{y}^{(0)}$ and $\hat{y}^{(1)}$ are 
    population-level contrasts rather than structural 
    counterfactuals for the same individual.
    
    \item \textbf{Distributional evaluability.} The model must 
    produce outputs compatible with our five reliability criteria 
    (subgroup calibration, tail variance ratio, arm 
    reconstruction error, HR recovery, and training stability) all of which require access to the full potential outcome 
    distributions $p(\hat{y}^{(a)})$ rather than point predictions.
\end{enumerate}

Table~\ref{tab:baseline_criteria} summarizes how each candidate 
model maps onto these criteria and the resulting inclusion decision.

\begin{table}[h]
\centering
\caption{Baseline selection against the three jointly necessary 
criteria. A model is included only if all three are satisfied.}
\label{tab:baseline_criteria}
\begin{tabular}{lcccc}
\toprule
Model & Generative & Invertible & Distributional & Included \\
      & mechanism  & abduction  & evaluability   &          \\
\midrule
CausalFlow-T (ours) & \cmark & \cmark & \cmark & \cmark \\
CVAE \citep{louizos2017} 
                    & \cmark & \xmark & \cmark & \cmark \\
GNN-CVAE            & \cmark & \xmark & \cmark & \cmark \\
TARNet \citep{shalit2017} 
                    & \xmark & \xmark & \xmark & \cmark \\
Causal Transformer \citep{melnychuk2022} 
                    & \xmark & \xmark & \xmark & \xmark \\
R-MSN \citep{lim2018}     
                    & \xmark & \xmark & \xmark & \xmark \\
CRN \citep{bica2020}      
                    & \xmark & \xmark & \xmark & \xmark \\
\bottomrule
\end{tabular}
\end{table}

Discriminative sequence models (including the Causal 
Transformer~\citep{melnychuk2022}, 
R-MSN~\citep{lim2018}, and 
CRN~\citep{bica2020}) fail all three criteria 
and are therefore excluded by construction, independently 
of their factual prediction performance.

We select CVAE~\citep{louizos2017}, GNN-CVAE, and 
TARNet~\citep{shalit2017} to span the key 
design axes underlying the proposed criteria. CVAE provides 
a variational inference baseline without causal structure; 
GNN-CVAE augments this with DAG-structured encoding under 
the same variational objective, isolating the contribution 
of causal structure within approximate inference. TARNet, 
in contrast, serves as a purely discriminative reference 
point, failing all three criteria. Together, these baselines expose the two failure modes:
confounding separation and structural propagation.

\subsection{Shared Components}
All models share: an outcome-reader LSTM ($h^Y_t$, used only in the encoder
to prevent outcome leakage), a treatment-encoder LSTM ($h^A_t$, replaced
from $t^*$ onwards at counterfactual generation time), and a covariate
aggregator (linear projection to $h^{\mathrm{cov}}_t$).
 
\subsection{Conditional CVAE}
\paragraph{Encoder.}
$q_\phi(z\mid x_{1:T},y_{1:T})=\mathcal{N}(\mu_\phi,\mathrm{diag}(\sigma^2_\phi))$,
where $[\mu_\phi,\log\sigma^2_\phi]=\mathrm{MLP}_\phi([\mathrm{flatten}(h^{\mathrm{cov}}_{1:T}),
\bar h^Y])$.  Latent dimension $L=64$.
 
\paragraph{Decoder.}
$\hat y_{1:T}=\mathrm{MLP}_\theta([z,\mathrm{flatten}(h^A_{1:T})])$.
 
\paragraph{Objective.}
$\mathcal{L}_{\mathrm{CVAE}}=-\mathbb{E}_{q_\phi}[\log
p_\theta(y_{1:T}\mid z,h^A_{1:T})]+\beta_t\,
\mathrm{KL}[q_\phi\|{\mathcal{N}(0,I)}]$,
with linear $\beta_t$ warm-up (0$\to$1 over first 20 epochs).
 
\paragraph{Key limitation.}
The variational posterior is an approximation; the abducted noise is not the
exact exogenous noise of the patient, and this approximation error propagates
into counterfactual predictions.
 
\subsection{GNN-CVAE}
Replaces the flat LSTM covariate encoder with a DAG-structured recurrent
encoder performing message passing in topological order:
\begin{equation}
  h^{(j)}_t = \mathrm{LayerNorm}\!\left(
    \mathrm{GRU}\!\left(h^{(j),\mathrm{in}}_t + m^{(j)}_t,\;
    h^{(j)}_{t-1}\right)\right),
\end{equation}
where $m^{(j)}_t=|{\mathrm{pa}(j)}|^{-1}\sum_{k\in\mathrm{pa}(j)}
\mathrm{MLP}_{k\to j}([h^{(k)}_t, h^{(j)}_{t-1}])$.
The causal structure is enforced in the encoder only. However the ELBO approximation gap in the abduction step remains, setting a ceiling that only exact inference can overcome.

\section{Baseline model for MNAR imputation: Conditional DAG-aware Flow Matching}
\label{app:CFM}

We implement a new Conditional DAG-aware Flow Matching (CFM) imputer as a
baseline for MNAR imputation of the synthetic
biomarker variables (SynthVars). CFM learns the conditional transport
$p_{\mathrm{noise}} \to p(\mathbf{x}^s_{\mathrm{miss}} \mid
\mathbf{x}^s_{\mathrm{obs}}, \mathbf{x}^{\mathrm{ns}}, T, Y, \tau)$
via flow matching, with a loss evaluated only at
missing positions.

\paragraph{Architecture.}
A Transformer encoder ($L{=}2$, $H{=}4$) processes
the always-observed non-SynthVar sequence $\mathbf{x}^{\mathrm{ns}}_{1:T}$
with a key-padding mask on unobserved timesteps, producing a leakage-free
context vector $\mathbf{h}_t$ that interpolates from truly-observed
neighbors on both sides of each gap.

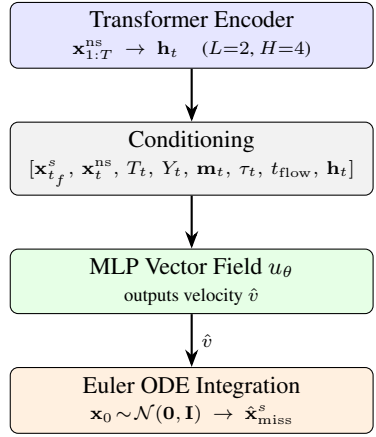
\begin{figure}[t]
\centering
\begin{tikzpicture}[
    every node/.style={font=\small},
    block/.style={draw, rounded corners=3pt,
                  minimum height=0.75cm, minimum width=4.8cm, align=center},
    arr/.style={->, thick, >=Stealth},
    node distance=0.7cm,
]
\node[block, fill=blue!10] (enc)
    {Transformer Encoder\\
     {\scriptsize $\mathbf{x}^{\mathrm{ns}}_{1:T}$ $\;\to\;$ $\mathbf{h}_t$
      \quad($L{=}2$, $H{=}4$)}};

\node[block, fill=gray!10, below=of enc] (cond)
    {Conditioning\\
     {\scriptsize $[\mathbf{x}^s_{t_f},\,\mathbf{x}^{\mathrm{ns}}_t,\,
       T_t,\,Y_t,\,\mathbf{m}_t,\,\tau_t,\,t_{\mathrm{flow}},\,\mathbf{h}_t]$}};

\node[block, fill=green!10, below=of cond] (mlp)
    {MLP Vector Field $u_\theta$\\
     {\scriptsize outputs velocity $\hat{v}$}};

\node[block, fill=orange!12, below=of mlp] (ode)
    {Euler ODE Integration\\
     {\scriptsize $\mathbf{x}_0\!\sim\!\mathcal{N}(\mathbf{0},\mathbf{I})$
      $\;\to\;$ $\hat{\mathbf{x}}^s_{\mathrm{miss}}$}};

\draw[arr] (enc)  -- (cond);
\draw[arr] (cond) -- (mlp);
\draw[arr] (mlp)  -- (ode) node[midway, right, font=\scriptsize] {$\hat{v}$};
\end{tikzpicture}
\caption{%
  \textbf{CFM imputer.}
  The Transformer Encoder produces context $\mathbf{h}_t$ from
  always-observed $\mathbf{x}^{\mathrm{ns}}_{1:T}$.
  The MLP $u_\theta$ is conditioned on $\mathbf{h}_t$ and DAG evidence
  ($T_t$, $Y_t$), trained with masked MSE on $v^*{=}\mathbf{x}_1{-}\mathbf{x}_0$.
  At inference, missing values are recovered by ODE integration from Gaussian noise.%
}
\label{fig:cfm_arch}
\end{figure}

A three-layer MLP vector field $u_\theta$ then maps the concatenation
\[
  \bigl[\mathbf{x}^s_{t_f} \;\|\; \mathbf{x}^{\mathrm{ns}}_t \;\|\;
  T_t \;\|\; Y_t \;\|\; \mathbf{m}_t \;\|\; \mathbf{x}^s_{t-1} \;\|\;
  \mathbf{m}_{t-1} \;\|\; \mathbf{h}_t \;\|\; \tau_t \;\|\;
  t_{\mathrm{flow}}\bigr]
\]
to a velocity $\hat{v} \in \mathbb{R}^{|\mathcal{S}|}$. Conditioning on $Y_t$
and $T_t$ ensures that downstream DAG evidence propagates into every imputed
value. The architecture is illustrated in Figure~\ref{fig:cfm_arch}.

\paragraph{Training.}
On fully-observed rows a random binary mask $\mathbf{m}^{\mathrm{miss}}$ is
applied and a linear interpolant
$\mathbf{x}_{t_f} = (1-t_f)\mathbf{x}_0 + t_f\mathbf{x}_1$,
$\mathbf{x}_0 \sim \mathcal{N}(\mathbf{0},\mathbf{I})$ is constructed, with
target velocity $v^* = \mathbf{x}_1 - \mathbf{x}_0$. The objective is a
masked mean-squared error at missing positions only, requiring no Jacobian:
\[
  \mathcal{L}(\theta)
  = \frac{1}{\|\mathbf{m}^{\mathrm{miss}}\|_1}
    \sum_{i} m^{\mathrm{miss}}_i
    \bigl(\hat{v}_i - v^*_i\bigr)^2.
\]

\paragraph{Inference.}
The encoder runs once over the full dataset to produce $\mathbf{h}_{1:T}$.
Missing values are then recovered by Euler integration of the ordinary
differential equation (ODE)
$\mathrm{d}\mathbf{x}/\mathrm{d}t_f =
u_\theta(\mathbf{x}_{t_f},\mathbf{c}_t,t_f)$
from Gaussian noise to the imputed target, with observed dimensions held
fixed throughout integration.

\section{Evaluation Protocol}
\label{app:eval}

\subsection{CausalFlow-T Evaluation Protocol}  
\label{app:eval_causal}

\paragraph{(i) Subgroup calibration.}
\[
  \mathrm{MAE}_q = \frac{1}{T}\sum_t\bigl|\widehat{\mathrm{ATE}}_q(t)
    -\mathrm{ATE}_q(t)\bigr|, \quad
  \mathrm{Bias}_q = \frac{1}{T}\sum_t\bigl(\widehat{\mathrm{ATE}}_q(t)
    -\mathrm{ATE}_q(t)\bigr),
\]
where $\widehat{\mathrm{ATE}}_q(t)=\mathbb{E}_{i\in q}[\hat{y}^{(1)}_i(t)]
-\mathbb{E}_{i\in q}[\hat{y}^{(0)}_i(t)]$ and analogously for the ground truth.
Quartiles are defined by the per-patient mean true ITE,
$\bar\tau_i = \frac{1}{T}\sum_t[y^{(1)}_i(t)-y^{(0)}_i(t)]$,
ensuring identical patient subgroups across all models.

\paragraph{(ii) Arm reconstruction error.}
\[
  \mathrm{Err}_a = \frac{1}{T}\sum_t
  \bigl|\mathbb{E}[\hat{y}^{(a)}(t)] - \mathbb{E}[y^{(a)}(t)]\bigr|,
  \quad a\in\{0,1\},
\]
measured in original outcome units, matching the scale of the mean trajectory plots.

\paragraph{(iii) Variance calibration.}
$\mathrm{VR}_{\mathrm{Q4}}=\mathrm{Var}(\hat{y}^{(a)}_{\mathrm{Q4}})/
\mathrm{Var}(y^{(a)}_{\mathrm{Q4}})$,
averaged over both arms and all time steps within Q4.
Values below~1 indicate distributional collapse (the model predicts near-constant
outputs for high-effect patients); values above~1 indicate overdispersion.






\subsection{Imputer Evaluation Protocol}
\label{app:eval_llm}

Let $\mathcal{M} = \{(i,t,j) : m_{i,t,j}=1\}$ denote the set of MNAR-missing biomarker cells, $\hat{x}_{i,t,j}$ the imputed value, and $x^{*}_{i,t,j}$ the ground truth (available by construction on the FIRE oracle, Section~\ref{sec:semi_synthetic}). Section~\ref{sec:exp:imputation} reports two layers: \emph{biomarker quality} (i)--(v) on the reconstruction itself, and \emph{downstream causal quality} (vi)--(ix) on the fixed CausalFlow-T estimator (Section~\ref{sec:exp:selection}). Throughout, $d{=}10$ is the number of synthetic biomarkers, $N$ the cohort size, and $T$ the number of timesteps.

\subsubsection{Biomarker quality}
\label{app:biomarker_quality}

\paragraph{(i) Pointwise MAE.}
\[
  \mathrm{MAE} = \frac{1}{|\mathcal{M}|}\sum_{(i,t,j)\in\mathcal{M}}
  \bigl| \hat{x}_{i,t,j} - x^{*}_{i,t,j} \bigr|,
\]
in original biomarker units, restricted to masked positions.

\paragraph{(ii) Pointwise RMSE.}
\[
  \mathrm{RMSE} =
  \sqrt{
  \frac{1}{|\mathcal{M}|}\sum_{(i,t,j)\in\mathcal{M}}
  \bigl( \hat{x}_{i,t,j} - x^{*}_{i,t,j} \bigr)^2 }.
\]
Like MAE, RMSE is computed in original biomarker units and restricted to
masked positions.

\paragraph{(iii) Lag-1 autocorrelation error.}
For biomarker $j$ and patient $i$, let
$r_{i,j}(z) = \mathrm{Corr}_t\bigl(z_{i,t,j},\, z_{i,t+1,j}\bigr)$
denote the Pearson correlation between consecutive timesteps of a series $z$. The mean lag-1 autocorrelations on the imputed and ground-truth datasets are,
\[
  \mathrm{AC} = \frac{1}{Nd}\sum_{i,j} r_{i,j}(\hat{x}),
  \qquad
  \mathrm{AC}^{*} = \frac{1}{Nd}\sum_{i,j} r_{i,j}(x^{*}),
\]
and we report $|\mathrm{AC}-\mathrm{AC}^{*}|$, with $\mathrm{AC}^{*}=0.366$. LOCF inflates AC toward~1; shrinkage-to-mean imputers deflate it toward~0.

\paragraph{(iv) Consecutive-step error.}
On observed-to-missing transitions $\mathcal{T} = \{(i,t,j) : m_{i,t,j}=0,\; m_{i,t+1,j}=1\}$,
\[
  \mathrm{CSE} = \frac{1}{|\mathcal{T}|}\sum_{(i,t,j)\in\mathcal{T}}
  \bigl| \hat{x}_{i,t+1,j} - x^{*}_{i,t+1,j} \bigr|,
\]
isolating one-step extrapolation from an observed clinical anchor.

\paragraph{(v) Biomarker--outcome correlation inflation.}
For each biomarker $j$, let
$\kappa_j(z) = \mathrm{Corr}\bigl(z_{\cdot,\cdot,j},\, y_{\cdot,\cdot}\bigr)$
be the Pearson correlation between the (imputed or true) biomarker series and the outcome, pooled over all $(i,t)$ pairs. We define
\[
  \mathrm{inf}_Y = \frac{1}{d}\sum_{j=1}^{d}
  \bigl(\kappa_j(\hat{x}) - \kappa_j(x^{*})\bigr),
\]
with $\mathrm{inf}_Y^{*}=0$. Positive values inflate the biomarker--outcome association, while negative values indicate deflation, in which a real signal is washed out.

\subsubsection{Downstream causal quality}
\label{app:downstream_causal_quality}

CausalFlow-T is held fixed (Section~\ref{sec:exp:selection}) so reported differences reflect imputation alone.

\paragraph{(vi) Subgroup calibration MAE.}
Following Appendix~\ref{app:eval}~(i), for each true-ITE quartile
$q \in \{Q_1, Q_2, Q_3, Q_4\}$ we compute
\[
  \mathrm{MAE}_{q} = \frac{1}{T}\sum_{t=1}^{T}
  \bigl|\widehat{\mathrm{ATE}}_{q}(t) - \mathrm{ATE}_{q}(t)\bigr|.
\]
We focus on the two tail quartiles, which expose complementary failure
modes invisible in aggregate metrics: $\mathrm{MAE}_{Q_1}$ on
low-effect patients, where confounding separation is hardest and
systematic bias is most exposed; and $\mathrm{MAE}_{Q_4}$ on
high-effect patients, where variance collapse manifests as flattening
of predicted treatment effects. Table~\ref{tab:imp_ranks} ranks
methods by $\mathrm{MAE}_{Q_1}$; Appendix
Table~\ref{tab:imp_causal} reports both tail quartiles.

\paragraph{(vii) Mean per-arm reconstruction error $\bar{\mathrm{Err}}_{a}$.}
As in Appendix~\ref{app:eval}~(ii), averaged across the two arms,
$\bar{\mathrm{Err}}_{a} = \tfrac{1}{2}\bigl(\mathrm{Err}_{a=0} +
\mathrm{Err}_{a=1}\bigr)$, summarizing population-level trajectory
accuracy under both treatment assignments.

\paragraph{(viii) Variance calibration.}
$\mathrm{VR}_{\mathrm{Q4}}$ as defined in Appendix~\ref{app:eval}~(iii).

\paragraph{(ix) Absolute ATE residual.} The ATE is the population-level expected difference between the potential outcome under treatment and under control, averaged across patients and timesteps:
\[
  \widehat{\mathrm{ATE}} = \frac{1}{NT}\sum_{i,t}
  \bigl(\hat{y}^{(1)}_i(t) - \hat{y}^{(0)}_i(t)\bigr),
  \qquad
  \mathrm{ATE}^{*} = \frac{1}{NT}\sum_{i,t}
  \bigl(y^{(1)}_i(t) - y^{(0)}_i(t)\bigr),
\]
with $\mathrm{ATE}^{*}=-3.484$. We report $|\widehat{\mathrm{ATE}}-\mathrm{ATE}^{*}|$.

\section{Training and Hyperparameter Details}
\label{app:training}
 Tables~\ref{tab:hyperparams} and~\ref{tab:llm_hparams} define the hyperparameter details for all the experimental parts.
 
\begin{table}[ht]
\centering
\caption{Hyperparameters across datasets}
\label{tab:hyperparams}
\resizebox{\textwidth}{!}{%
\begin{tabular}{llccc}
\toprule
\textbf{Component} & \textbf{Hyperparameter} & \textbf{Synthetic} & \textbf{FIRE semi} & \textbf{FIRE real} \\
\midrule
\multirow{5}{*}{Shared training}
  & Optimizer          & Adam        & Adam & Adam \\
  & Learning rate      & $10^{-3}$   & $10^{-3}$ & $10^{-4}$ \\
  & Batch size         & 512         & 256 & 64 \\
  & Data split (train(train/val)/test) & 80(80/20)/20 & 80(80/20)/20 & 80(80/20)/20 \\
\midrule
\multirow{2}{*}{LSTM encoder}
  & Hidden dimension $H$ & 128       & 256 & 256 \\
  & Dropout              & -       & - & 0.3 \\
\midrule
\multirow{2}{*}{CausalMAF}
  & Coupling layers    & 1           & 1 & 1 \\
  & Hidden dimension   & 128         & 128 & 128 \\
\midrule
\multirow{2}{*}{TARNet}
  & Hidden dimension   & 128         & 128 & Not Used \\
  & MLP depth          & 2           & 1   & Not Used \\
\midrule
\multirow{2}{*}{GNN-CVAE}
  & Encoder layers & 2           & 2 & Not Used \\
  & Latent dimension   & 64          & 64 & Not Used \\
\midrule
CVAE
  & Architecture       & No graph encoder & Not Used & Not Used \\
\midrule
\multirow{2}{*}{Normalization}
  & Covariates \& treatments & Joint standardization & Joint standardization & Per-timestep \\
  & Outcomes           & Per-timestep & Per-timestep & Per-timestep \\
\bottomrule
\end{tabular}}
\end{table}

\begin{table}[ht]
\centering
\caption{Main hyperparameters of the LLM-driven evolutionary imputation search.}
\label{tab:llm_hparams}
\begin{tabular}{ll}
\toprule
\textbf{Hyperparameter} & \textbf{Value} \\
\midrule
LLM model                                 & \texttt{gpt-5.4, qwen3.5-plus, gpt-oss-120b} \\
Iteration budget $K$                      & 20 \\
Proxy holdout fraction $\rho$             & 0.10 \\
Holdout base seed $\sigma$                & fixed across candidate evaluations  \\
Composite-score weight $\lambda_Y$        & 2.0 \\
Composite-score weight $\lambda_T$        & 0.5 \\
Selection rule                            & minimize $s = \mathrm{RMSE} + \lambda_Y\,\Delta_Y + \lambda_T\,\Delta_T$ \\
History window $W$                        & 3 \\
Per-dataset timeout $\tau_{\mathrm{run}}$ & 180\,s \\
Memory cap $M_{\mathrm{run}}$             & 6144\,MB \\
\bottomrule
\end{tabular}
\end{table}

\section{FIRE Semi-Synthetic Oracle Results}
\label{app:fire_oracle}
Table~\ref{tab:fire_oracle} reports full results on the FIRE semi-synthetic
oracle (no missingness), used as a controlled bridge between synthetic
benchmarks and the imputation experiments
(Section~\ref{sec:exp:imputation}).
True $\mathrm{ATE}\approx{-3.48}$; $n{=}10$ seeds throughout.

\begin{table}[ht]
\centering
\footnotesize
\setlength{\tabcolsep}{2pt}
\caption{FIRE semi-synthetic oracle: systematic-error ratio
$|\mathrm{Bias_{Q1}}|/\mathrm{MAE_{Q1}}$ ($\downarrow$;
${<}0.5$ = predominantly random errors),
arm reconstruction, variance ratio, and ATE recovery ($n{=}10$).
$\mathrm{VR}_{\mathrm{Q4}}\!\to\!1$.
True $\mathrm{ATE}\approx{-3.48}$.
\textbf{Bold}: best per metric.}
\label{tab:fire_oracle}
\resizebox{\linewidth}{!}{%
\begin{tabular}{l|c|cc|c|c}
\toprule
\textbf{Model}
  & $|\mathrm{Bias_{Q1}}|/\mathrm{MAE_{Q1}}{\downarrow}$
  & ${\mathrm{Err}}_{a=1}$ 
  & ${\mathrm{Err}}_{a=0}$
  & $\mathrm{VR}_{\mathrm{Q4}}$
  & $\mathrm{ATE}$\ \\
\midrule
CausalFlow-T \textbf{(ours)}
  & \textbf{0.316}
  & 0.088$\pm$0.032 & \textbf{0.063$\pm$0.022}
  & \textbf{1.006$\pm$0.009}
  & $-$3.568$\pm$0.112 \\
NF (no DAG)
  & 0.883
  & 0.161$\pm$0.073 & 0.084$\pm$0.044
  & 1.016$\pm$0.010
  & $-$3.382$\pm$0.251 \\
TARNet
  & 0.708
  & \textbf{0.081$\pm$0.012} & 0.078$\pm$0.003
  & 0.608$\pm$0.003
  & \textbf{$-$3.415$\pm$0.015} \\
GNN-CVAE (DAG)
  & 0.351
  & 0.216$\pm$0.048 & 0.202$\pm$0.038
  & 0.616$\pm$0.021
  & $-$3.395$\pm$0.113 \\
CVAE
  & 0.938
  & 0.134$\pm$0.045 & 0.127$\pm$0.047
  & 0.482$\pm$0.055
  & $-$3.451$\pm$0.075 \\
\bottomrule
\end{tabular}
}
\vspace{4pt}
\end{table}

CausalFlow-T ($0.316$) and GNN-CVAE ($0.351$) are the only two models
passing the ${<}0.5$ reliability threshold, indicating that
low-responder errors are predominantly random rather than systematic;
NF (no DAG) ($0.883$), TARNet ($0.708$), and CVAE ($0.938$) all
fail, confirming that explicit causal structure is necessary for
confounding separation under realistic clinical covariate distributions.
CausalFlow-T is further distinguished as the only model with
near-perfect variance preservation ($\mathrm{VR}\!=\!1.006\pm0.009$),
while GNN-CVAE collapses ($\mathrm{VR}\!=\!0.616$) and CVAE collapses
more severely ($\mathrm{VR}\!=\!0.482$), confirming that passing the
bias threshold alone is insufficient for individual-level reliability.
TARNet achieves the best arm-1 reconstruction ($0.081$) and closest
ATE point estimate ($-3.415$) but fails the ratio threshold ($0.708$)
and collapses variance ($\mathrm{VR}\!=\!0.608$), reinforcing that
factual accuracy does not imply counterfactual reliability.
Removing the DAG constraint (NF no DAG) degrades the bias ratio to
$0.883$ and produces the largest ATE deviation ($-3.382$,
$\Delta\!=\!0.102$) with substantial cross-seed instability
($\pm0.251$), underscoring that structural supervision is necessary
for stable estimation under clinical covariate distributions.

\section{Full Numerical Results}
\label{app:full_results}

Tables~\ref{tab:calibration} and~\ref{tab:structural} report complete
per-model, per-benchmark results for the four synthetic benchmarks;
they are the primary evidence base for Table~\ref{tab:ranks} in the
main body. FIRE semi-synthetic oracle results are in
Table~\ref{tab:fire_oracle} (Appendix~\ref{app:fire_oracle}).

\begin{itemize}

\item \textbf{Simple 3-Node.}
      On \textit{Simple 3-Node} (linear, no confounding), all models
      show $|\mathrm{Bias}_q| \approx \mathrm{MAE}_q$ with near-zero
      standard deviations throughout.
      This is expected: in a confounding-free linear setting, residual
      errors are purely systematic rather than variance-driven.
      Seed consistency confirms training stability across all
      architectures, so differences on harder benchmarks reflect genuine
      architectural sensitivity.
      GNN-CVAE achieves the lowest Q1 MAE ($0.425$) in this regime,
      illustrating that explicit graph encoding can reduce absolute
      error when confounding is absent (but this advantage disappears
      under the more complex generating processes below).

\item \textbf{LDL Toy.}
      TARNet achieves the lowest absolute MAE across all quartiles
      ($0.392$--$2.107$), but maintains a
      $|\mathrm{Bias}_q|/\mathrm{MAE}_q$ ratio at or near $1.0$
      throughout (every patient in a given quartile is wrong in the
      same direction), indicating systematic confounding failure rather
      than calibrated uncertainty.
      CausalFlow-T achieves $\mathrm{VR}_{\mathrm{Q4}} = 1.049 \pm
      0.033$ (closest to $1$ across all models), confirming that tail
      variance is faithfully preserved rather than collapsed or inflated,
      while maintaining sign-consistent positive bias
      ($+0.547$--$+0.963$ across quartiles) indicating mild but coherent
      overestimation rather than directionally inconsistent errors.
      NF\,(no~DAG) systematically underestimates across all quartiles
      ($-0.402$ to $-0.604$), confirming that causal factorization
      is necessary for directionally correct inference.
      GNN-CVAE and CVAE show variance collapse
      ($\mathrm{VR}_{\mathrm{Q4}} \leq 0.85$) and substantially larger
      arm errors, revealing distributional narrowing in the
      high-effect tail.

\item \textbf{CVD Risk}
      Ground-truth $\mathrm{HR} = 0.831$ (protective treatment).
      CausalFlow-T recovers $\mathrm{HR} = 0.786 \pm 0.051$
      ($\Delta = 0.045$), the only model within clinically meaningful
      range with arm errors an order of magnitude below all competitors.
      GNN-CVAE and CVAE collapse to $\mathrm{HR} \approx 1.005 \pm
      0.030$ across all seeds (predicting a null effect despite a
      true 17\% hazard reduction) consistent with posterior collapse
      under the ELBO objective.
      TARNet predicts $\mathrm{HR} = 1.133 \pm 0.148$, inverting the
      effect direction on average.
      NF\,(no~DAG) yields $\mathrm{HR} = 0.834 \pm 0.324$: the large
      standard deviation (related to the explosion in the variance inf Q4) reveals that without the DAG constraint the
      flow is fundamentally unstable under strong mediation
      (sometimes approximately correct, sometimes catastrophically wrong),
      making it unreliable for clinical deployment regardless of
      average performance.

\item \textbf{Cox Survival.}
      CVAE achieves lower Q1--Q4 MAE on Cox ($0.023 \pm 0.001$--$0.020
      \pm 0.006$) because variational models handle binary outcomes
      without continuous relaxation.
      However, CausalFlow-T achieves the best arm-1 reconstruction
      error ($0.014 \pm 0.002$) and closest HR recovery
      ($0.866 \pm 0.011$ vs.\ true $0.887$), while CVAE and GNN-CVAE
      show arm errors $4$--$6\times$ higher.
      TARNet achieves the best arm-0 error ($0.005 \pm 0.001$) but
      overshoots the HR ($0.966 \pm 0.005$), indicating factual
      accuracy without causal calibration.
      The Cox MAE advantage of CVAE does not generalize to more complex
      DAGs, as the CVD Risk results confirm.

\end{itemize}

\paragraph{CVAE vs.\ GNN-CVAE.}
Across benchmarks, CVAE consistently outperforms GNN-CVAE despite the
latter incorporating explicit graph structure in the encoder.
This contrasts with the normalizing flow setting, where the DAG
constraint yields consistent improvement.
Both CVAE variants rely on variational inference: the ELBO
approximation introduces an irreducible inference gap that dominates
performance, limiting the benefit of improved structural inductive
bias.
The GNN encoder does not translate into better counterfactual estimates
and introduces additional optimization complexity, evidenced by
GNN-CVAE's substantially higher LDL arm-1 error ($4.854 \pm 0.451$)
relative to plain CVAE ($0.535 \pm 0.245$).
Normalizing flows perform exact likelihood-based inference, so
architectural improvements such as DAG factorization directly improve
abduction quality.
This explains the pattern CVAE $>$ GNN-CVAE, whereas
CausalFlow-T $>$ NF\,(no~DAG).

\section{Full Imputation Results}
\label{app:imp_results}

\begin{figure}[ht]
  \centering
  \includegraphics[width=0.95\linewidth]{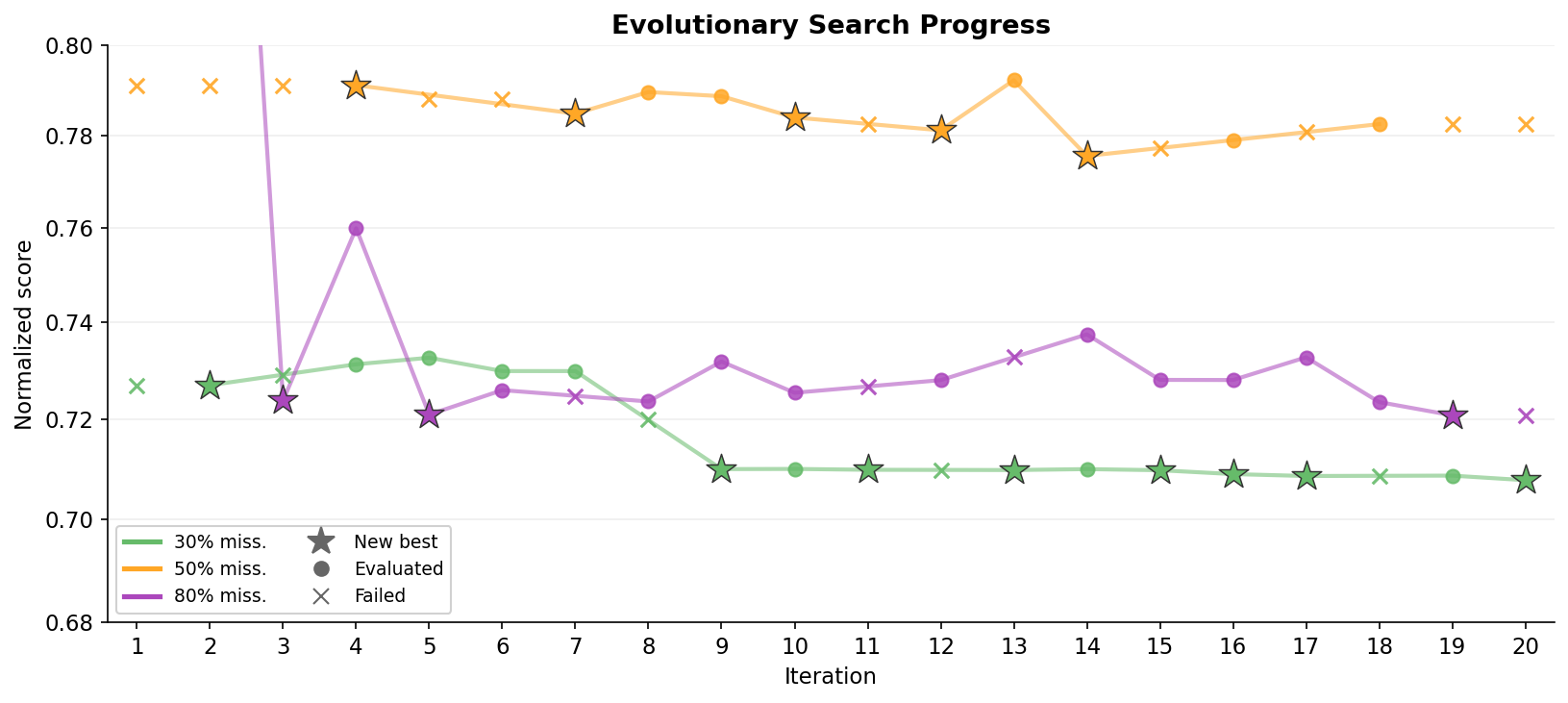}
  \caption{Evolutionary search progress for the LLM-driven imputer across 30\%, 50\%, and 80\% MNAR missingness, shown as normalized proxy scores over the 20-iteration budget.}
  \label{fig:llm_search_progress}
\end{figure}

Tables~\ref{tab:biomarker_quality} and~\ref{tab:imp_causal} report
complete per-method, per-missingness-level results underlying
Table~\ref{tab:imp_ranks}. Figure~\ref{fig:llm_search_progress}
summarizes the normalized proxy-score trajectories during the selected
GPT-5.4 evolutionary search, showing rapid early improvement followed by
smaller accepted refinements.

To separate the effect of the evolutionary loop from a single executable
LLM proposal, the tables also report ``GPT-5.4 first-valid'', defined as
the first GPT-5.4 candidate that passed static and runtime checks and was
evaluated. This row is diagnostic only and is not included in the pooled
ranks used for model selection. The first-valid rows show that executable
LLM proposals are not sufficient on their own, especially at 50\%--80\%
missingness, where the final selected imputer substantially improves
biomarker reconstruction and ATE recovery.

Biomarker-level metrics are computed once over all imputed cells across
the ten semi-synthetic biomarkers and are therefore reported without seed
variability; downstream causal metrics are averaged over $n=10$ training
seeds.

\begin{table}[ht]
\centering
\footnotesize
\setlength{\tabcolsep}{4pt}
\caption{Biomarker-level imputation quality across missingness levels,
averaged over the semi-synthetic biomarkers. Pointwise MAE and RMSE are
computed at missing positions only in original biomarker units.
Lag-1 autocorrelation $\mathrm{AC}$ is compared against the ground-truth
value $\mathrm{AC}^{*}=0.375$; CSE denotes consecutive-step error; and
$\mathrm{inf}_Y$ is biomarker--outcome correlation inflation
(ground truth: $0$).``GPT-5.4 first-valid'' denotes the first executable GPT-5.4 proposal
evaluated during the evolutionary search.
} 
\label{tab:biomarker_quality}
\resizebox{\linewidth}{!}{%
\begin{tabular}{ll|ccccc}
\toprule
\textbf{Missing} & \textbf{Method}
  & MAE $\downarrow$
  & RMSE $\downarrow$
  & $\mathrm{AC}$ ($\to 0.375$)
  & CSE $\downarrow$
  & $\mathrm{inf}_Y$ ($\to 0$) \\
\midrule
\multirow{6}{*}{30\%}
  & LOCF             & 3.314 & 4.399 & 0.473 & 2.781 & $-0.113$ \\
  & MissForest       & 2.090 & 2.674 & 0.384 & 2.073 & $+0.036$ \\
  & CausalCFM        & 2.755 & 3.523 & 0.350 & 2.708 & $-0.007$ \\
  & GPT-5.4      & \textbf{1.914} & \textbf{2.461} & 0.392 & \textbf{1.893} & $+0.034$ \\
  & Qwen3.5-Plus & 2.055 & 2.659 & \textbf{0.375} & 1.956 & \textbf{$+0.003$} \\
  & GPT-OSS-120b & 1.972 & 2.548 & 0.386 & 1.914 & $+0.026$ \\
    \addlinespace[1pt]
  \cmidrule(lr){2-7}
  \addlinespace[1pt]
  & GPT-5.4 first-valid    & 2.026 & 2.598 & 0.389 & 2.009 & $+0.036$ \\

\midrule
\multirow{6}{*}{50\%}
  & LOCF             & 3.378 & 4.463 & 0.554 & 2.773 & $-0.197$ \\
  & MissForest       & 2.219 & 2.839 & \textbf{0.390} & 2.185 & $+0.059$ \\
  & CausalCFM        & 2.833 & 3.623 & 0.334 & 2.758 & \textbf{$-0.009$} \\
  & GPT-5.4      & \textbf{2.052} & \textbf{2.628} & 0.413 & \textbf{2.003} & $+0.052$ \\
  & Qwen3.5-Plus & 2.461 & 3.163 & 0.314 & 2.349 & $-0.013$ \\
  & GPT-OSS-120b & 2.059 & 2.646 & 0.423 & 2.011 & $+0.054$ \\
    \addlinespace[1pt]
  \cmidrule(lr){2-7}
  \addlinespace[1pt]
  & GPT-5.4 first-valid   & 2.805 & 3.741 & 0.619 & 2.343 & $-0.138$ \\

\midrule
\multirow{6}{*}{80\%}
  & LOCF             & 3.407 & 4.509 & 0.713 & 2.760 & $-0.347$ \\
  & MissForest       & 2.504 & 3.187 & \textbf{0.380} & 2.468 & $+0.072$ \\
  & CausalCFM        & 2.875 & 3.679 & 0.348 & 2.798 & \textbf{$-0.011$} \\
  & GPT-5.4      & \textbf{2.059} & \textbf{2.634} & 0.596 & \textbf{2.035} & $+0.159$ \\
  & Qwen3.5-Plus & 2.788 & 3.645 & 0.208 & 2.612 & $-0.080$ \\
  & GPT-OSS-120b & 2.147 & 2.745 & 0.514 & 2.109 & $+0.129$ \\
    \addlinespace[1pt]
  \cmidrule(lr){2-7}
  \addlinespace[1pt]
  & GPT-5.4 first-valid    & 3.146 & 4.193 & 0.729 & 2.442 & $-0.318$ \\

\bottomrule
\end{tabular}
}
\end{table}

\begin{table}[ht]
\centering
\footnotesize
\setlength{\tabcolsep}{4pt}
\caption{Downstream causal inference quality. All methods use the same
fixed CausalFlow-T estimator (Section~\ref{sec:exp:selection}); only
the imputation strategy varies. Q1 MAE and Q4 MAE (lowest- and
highest-ITE quartiles), mean per-arm reconstruction error
$\bar{\mathrm{Err}}_{a}=\tfrac{1}{2}(\mathrm{Err}_{a=0}+\mathrm{Err}_{a=1})$,
tail variance ratio $\mathrm{VR}_{\mathrm{Q4}}$ (target value $1$),
and absolute ATE residual against $\mathrm{ATE}^{*} = -3.484$.
Means $\pm$ standard deviations over $n=10$ seeds.
$\dagger$~Oracle reference: CausalFlow-T on fully observed data,
$\bar{\mathrm{Err}}_{a} = 0.058$ (Table~\ref{tab:fire_oracle}). ``GPT-5.4 first-valid'' denotes the first executable GPT-5.4 proposal evaluated during the evolutionary search.}
\label{tab:imp_causal}
\begin{tabular}{ll|cc|c|c|c}
\toprule
\textbf{Missing} & \textbf{Method}
  & Q1 MAE $\downarrow$ & Q4 MAE $\downarrow$
  & $\bar{\mathrm{Err}}_{a}$ $\downarrow$
  & $\mathrm{VR}_{\mathrm{Q4}}$
  & $|\Delta \mathrm{ATE}|$ $\downarrow$ \\
\midrule
\multirow{4}{*}{30\%}
  & LOCF        & 0.224 $\pm$ 0.123 & 0.153 $\pm$ 0.022 & 0.079 $\pm$ 0.023 & 1.005 $\pm$ 0.006 & 0.090 $\pm$ 0.113 \\
  & MissForest  & 0.416 $\pm$ 0.137 & 0.401 $\pm$ 0.144 & 0.177 $\pm$ 0.093 & 0.998 $\pm$ 0.003 & 0.331 $\pm$ 0.223 \\
  & CausalCFM   & \textbf{0.161 $\pm$ 0.073$^{\dagger}$} & \textbf{0.118 $\pm$ 0.015} & \textbf{0.059 $\pm$ 0.006} & 1.008 $\pm$ 0.005 & 0.027 $\pm$ 0.066 \\
  & GPT-5.4 & 0.223 $\pm$ 0.120 & 0.159 $\pm$ 0.036 & 0.071 $\pm$ 0.016 & 1.007 $\pm$ 0.011 & \textbf{0.024 $\pm$ 0.120} \\
  & Qwen3.5-Plus    & 0.201 $\pm$ 0.107 & 0.220 $\pm$ 0.054 & 0.086 $\pm$ 0.035 & 1.004 $\pm$ 0.004 & 0.145 $\pm$ 0.092 \\
  & GPT-OSS-120b & 0.180 $\pm$ 0.040 & 0.181 $\pm$ 0.034 & 0.065 $\pm$ 0.010 & 1.005 $\pm$ 0.007 & 0.075 $\pm$ 0.053 \\
  \addlinespace[1pt]
  \cmidrule(lr){2-7}
  \addlinespace[1pt]
    & GPT-5.4 first-valid       & 0.363 $\pm$ 0.239 & 0.556 $\pm$ 0.159 & 0.233 $\pm$ 0.073 & 1.008 $\pm$ 0.009 & 0.464 $\pm$ 0.199 \\

\midrule
\multirow{4}{*}{50\%}
  & LOCF        & 0.224 $\pm$ 0.074 & 0.146 $\pm$ 0.049 & 0.084 $\pm$ 0.031 & 1.008 $\pm$ 0.008 & \textbf{0.007 $\pm$ 0.174} \\
  & MissForest  & 0.305 $\pm$ 0.143 & 0.342 $\pm$ 0.192 & 0.155 $\pm$ 0.084 & 0.999 $\pm$ 0.006 & 0.184 $\pm$ 0.306 \\
  & CausalCFM   & 0.232 $\pm$ 0.105 & 0.206 $\pm$ 0.105 & 0.096 $\pm$ 0.048 & 1.007 $\pm$ 0.006 & 0.048 $\pm$ 0.198 \\
  & GPT-5.4 & 0.219 $\pm$ 0.067 & 0.199 $\pm$ 0.043 & 0.083 $\pm$ 0.029 & 1.012 $\pm$ 0.007 & 0.131 $\pm$ 0.085 \\
  & Qwen3.5-Plus    & \textbf{0.187 $\pm$ 0.099} & \textbf{0.132 $\pm$ 0.034} & \textbf{0.069 $\pm$ 0.022} & 1.008 $\pm$ 0.006 & 0.013 $\pm$ 0.114 \\
  & GPT-OSS-120b & 0.194 $\pm$ 0.095 & 0.283 $\pm$ 0.120 & 0.116 $\pm$ 0.056 & 1.010 $\pm$ 0.006 & 0.199 $\pm$ 0.149 \\
  \addlinespace[1pt]
  \cmidrule(lr){2-7}
  \addlinespace[1pt]
    & GPT-5.4 first-valid      & 0.263 $\pm$ 0.146 & 0.285 $\pm$ 0.191 & 0.129 $\pm$ 0.063 & 1.005 $\pm$ 0.008 & 0.216 $\pm$ 0.201 \\

\midrule
\multirow{4}{*}{80\%}
  & LOCF        & 0.235 $\pm$ 0.045 & \textbf{0.176 $\pm$ 0.050} & 0.088 $\pm$ 0.031 & 1.009 $\pm$ 0.005 & 0.052 $\pm$ 0.163 \\
  & MissForest  & \textbf{0.175 $\pm$ 0.044} & 0.253 $\pm$ 0.169 & 0.102 $\pm$ 0.070 & 1.001 $\pm$ 0.002 & 0.145 $\pm$ 0.188 \\
  & CausalCFM   & 0.548 $\pm$ 0.358 & 0.440 $\pm$ 0.244 & 0.226 $\pm$ 0.132 & 1.037 $\pm$ 0.024 & 0.408 $\pm$ 0.332 \\
  & GPT-5.4 & 0.311 $\pm$ 0.133 & 0.183 $\pm$ 0.086 & 0.093 $\pm$ 0.050 & 1.016 $\pm$ 0.005 & \textbf{0.013 $\pm$ 0.191} \\
  & Qwen3.5-Plus    & 0.370 $\pm$ 0.220 & 0.219 $\pm$ 0.080 & 0.133 $\pm$ 0.063 & 1.032 $\pm$ 0.007 & 0.190 $\pm$ 0.185 \\
  & GPT-OSS-120b & 0.222 $\pm$ 0.105 & 0.211 $\pm$ 0.153 & \textbf{0.088 $\pm$ 0.044} & 1.019 $\pm$ 0.008 & 0.072 $\pm$ 0.172 \\
  \addlinespace[1pt]
  \cmidrule(lr){2-7}
  \addlinespace[1pt]
    & GPT-5.4 first-valid     & 0.283 $\pm$ 0.150 & 0.327 $\pm$ 0.182 & 0.140 $\pm$ 0.064 & 1.010 $\pm$ 0.006 & 0.201 $\pm$ 0.205 \\
\bottomrule
\end{tabular}
\end{table}

\begin{sidewaystable}
\centering
\caption{Subgroup calibration and tail variance ratio across all benchmarks.
GNN-CVAE and CVAE on LDL and Cox use the full simulation variant.
\textbf{Bold}: best value(s) per dataset per metric. ``expl.'' = numerical explosion.}
\label{tab:calibration}
\resizebox{\textwidth}{!}{%
\begin{tabular}{ll|rrrr|rrrr|r}
\toprule
\textbf{Dataset} & \textbf{Model}
  & \multicolumn{4}{c|}{\textbf{Quartile MAE}}
  & \multicolumn{4}{c|}{\textbf{Quartile Bias}}
  & \textbf{VR\,Q4} \\
\cmidrule(lr){3-6}\cmidrule(lr){7-10}
& & Q1 & Q2 & Q3 & Q4 & Q1 & Q2 & Q3 & Q4 & \\
\midrule
Simple     & CausalFlow-T  & 0.533$\pm$0.002 & 0.127$\pm$0.001 & 0.151$\pm$0.005 & 0.553$\pm$0.001
           & 0.433$\pm$0.002 & 0.110$\pm$0.001 & $-$0.148$\pm$0.001 & \textbf{$-$0.353$\pm$0.001} & \textbf{0.991$\pm$0.001} \\
3-Node     & NF (no DAG)   & 0.499$\pm$0.000 & \textbf{0.093$\pm$0.000} & 0.182$\pm$0.000 & 0.586$\pm$0.000
           & 0.499$\pm$0.000 & 0.093$\pm$0.000 & $-$0.182$\pm$0.000 & $-$0.586$\pm$0.000 & \textbf{0.991$\pm$0.000} \\
           & GNN-CVAE      & \textbf{0.425$\pm$0.000} & 0.110$\pm$0.000 & 0.248$\pm$0.000 & 0.659$\pm$0.000
           & \textbf{0.425$\pm$0.000} & \textbf{0.028$\pm$0.000} & $-$0.248$\pm$0.000 & $-$0.659$\pm$0.000 & 0.852$\pm$0.000 \\
           & CVAE          & 0.514$\pm$0.000 & 0.121$\pm$0.000 & 0.156$\pm$0.000 & 0.566$\pm$0.000
           & 0.514$\pm$0.000 & 0.121$\pm$0.000 & $-$0.156$\pm$0.000 & $-$0.566$\pm$0.000 & 0.763$\pm$0.000 \\
           & TARNet        & 0.549$\pm$0.000 & 0.147$\pm$0.000 & \textbf{0.133$\pm$0.000} & \textbf{0.546$\pm$0.000}
           & 0.549$\pm$0.000 & 0.147$\pm$0.000 & \textbf{$-$0.133$\pm$0.000} & $-$0.546$\pm$0.000 & 0.810$\pm$0.000 \\
\midrule
LDL        & CausalFlow-T  & 2.029$\pm$0.201 & 2.326$\pm$0.289 & 2.393$\pm$0.449 & 3.564$\pm$0.626
           & 0.547$\pm$0.639 & \textbf{0.677$\pm$0.569} & \textbf{0.781$\pm$0.657} & \textbf{0.963$\pm$1.020} & \textbf{1.049$\pm$0.033} \\
           & NF (no DAG)   & 2.057$\pm$0.320 & 2.449$\pm$0.427 & 2.752$\pm$0.442 & 3.769$\pm$0.453
           & \textbf{$-$0.402$\pm$0.356} & $-$0.713$\pm$0.734 & \textbf{$-$0.773$\pm$0.641}& $-$1.004$\pm$0.650 & 1.098$\pm$0.025 \\
           & GNN-CVAE$^\dagger$ & 23.056$\pm$1.133 & 13.111$\pm$0.833 & 7.563$\pm$0.658 & \textbf{1.729$\pm$0.565}
           & 23.035$\pm$1.170 & 13.071$\pm$0.854 & 7.516$\pm$0.653 & 1.677$\pm$0.568 & 0.846$\pm$0.033 \\
           & CVAE$^\dagger$ & 7.005$\pm$0.554 & 1.540$\pm$0.242 & 1.477$\pm$0.418 & 4.216$\pm$0.499
           & 6.809$\pm$0.543 & 1.296$\pm$0.311 & $-$1.477$\pm$0.418 & $-$4.212$\pm$0.494 & 0.698$\pm$0.026 \\
           & TARNet        & \textbf{0.392$\pm$0.081} & \textbf{0.842$\pm$0.071} & \textbf{1.003$\pm$0.065} & 2.107$\pm$0.114
           & $-$0.546$\pm$0.126 & 0.751$\pm$0.074 & 0.885$\pm$0.068 & 2.107$\pm$0.114 & 0.639$\pm$0.003 \\
\midrule
Cox        & CausalFlow-T  & 0.025$\pm$0.008 & 0.025$\pm$0.003 & 0.027$\pm$0.003 & 0.034$\pm$0.008
           & \textbf{$-$0.005$\pm$0.018} & $-$0.010$\pm$0.008 & $-$0.011$\pm$0.007 & $-$0.021$\pm$0.013 & 1.528$\pm$0.016 \\
Survival   & NF (no DAG)   & \textbf{0.022$\pm$0.005} & 0.024$\pm$0.006 & 0.028$\pm$0.009 & 0.038$\pm$0.009
           & $-$0.011$\pm$0.011 & $-$0.012$\pm$0.013 & $-$0.016$\pm$0.015 & $-$0.029$\pm$0.012 & 1.522$\pm$0.009 \\
           & GNN-CVAE$^\dagger$ & 0.026$\pm$0.012 & 0.020$\pm$0.007 & 0.015$\pm$0.005 & \textbf{0.018$\pm$0.002}
           & $-$0.014$\pm$0.017 & $-$0.006$\pm$0.012 & \textbf{$-$0.002$\pm$0.007} & \textbf{$-$0.000$\pm$0.003} & \textbf{0.916$\pm$0.035} \\
           & CVAE$^\dagger$ & \textbf{0.023$\pm$0.001} & \textbf{0.014$\pm$0.007} & \textbf{0.014$\pm$0.007} & 0.020$\pm$0.006
           & 0.011$\pm$0.018 & \textbf{0.002$\pm$0.011} & $-$0.002$\pm$0.008 & $-$0.002$\pm$0.006 & 0.849$\pm$0.038 \\
           & TARNet        & 0.047$\pm$0.003 & 0.047$\pm$0.002 & 0.042$\pm$0.004 & 0.029$\pm$0.003
           & 0.047$\pm$0.003 & 0.047$\pm$0.002 & 0.042$\pm$0.004 & 0.029$\pm$0.003 & 0.855$\pm$0.058 \\
\midrule
CVD Risk   & CausalFlow-T  & \textbf{0.010$\pm$0.005} & \textbf{0.007$\pm$0.004} & \textbf{0.004$\pm$0.002} & \textbf{0.003$\pm$0.001}
           & \textbf{$-$0.007$\pm$0.008} & \textbf{$-$0.005$\pm$0.005} & \textbf{$-$0.003$\pm$0.003} & \textbf{$0.000\pm$0.002} & \textbf{1.036$\pm$0.001} \\
Toy        & NF (no DAG)   & 0.023$\pm$0.010 & 0.009$\pm$0.004 & 0.006$\pm$0.003 & 0.557$\pm$1.750
           & 0.021$\pm$0.011 & 0.007$\pm$0.007 & 0.003$\pm$0.007 & 0.555$\pm$1.751& expl. \\
           & GNN-CVAE      & 0.027$\pm$0.001 & 0.011$\pm$0.000 & 0.008$\pm$0.000 & 0.004$\pm$0.000
           & 0.027$\pm$0.001 & 0.011$\pm$0.000 & 0.008$\pm$0.000 & 0.004$\pm$0.000 & expl. \\
           & CVAE          & 0.027$\pm$0.000 & 0.012$\pm$0.000 & 0.008$\pm$0.000 & 0.004$\pm$0.000
           & 0.027$\pm$0.000 & 0.012$\pm$0.000 & 0.008$\pm$0.000 & 0.004$\pm$0.000 & expl. \\
           & TARNet        & 0.080$\pm$0.011 & 0.022$\pm$0.011 & 0.011$\pm$0.007 & 0.013$\pm$0.010
           & 0.080$\pm$0.011 & 0.021$\pm$0.012 & 0.001$\pm$0.013 & $-$0.007$\pm$0.014 & \textbf{1.032$\pm$0.051} \\
\bottomrule
\multicolumn{11}{l}{$^\dagger$ GNN-CVAE and CVAE evaluated on the full LDL / full Cox simulation variant.}
\end{tabular}
}
\end{sidewaystable}

\begin{table*}[t]
\centering
\footnotesize
\setlength{\tabcolsep}{4pt}
\caption{Arm reconstruction errors and ATE/HR recovery.
\textbf{Bold}: best per dataset per metric.
$^\dagger$ GNN-CVAE and CVAE on LDL/Cox use the full simulation variant.}
\label{tab:structural}
\begin{tabular}{ll|cc|c}
\toprule  
\textbf{Dataset} & \textbf{Model}
    & ${\mathrm{Err}}_{a=1}$ 
  & ${\mathrm{Err}}_{a=0}$
  & $\mathrm{ATE}$  or $\mathrm{HR}$  (true\,/\,pred) \\
\midrule
Simple     & CausalFlow-T  & \textbf{0.025$\pm$0.004} & \textbf{0.019$\pm$0.003} & $-$0.940\,/\,$-$0.950$\pm$0.001 \\
3-Node     & NF (no DAG)   & 0.028$\pm$0.000 & 0.022$\pm$0.000 & $-$0.940\,/\,$-$0.984$\pm$0.000 \\
           & GNN-CVAE      & 0.203$\pm$0.000 & 0.239$\pm$0.000 & $-$0.942\,/\,$-$1.055$\pm$0.000 \\
           & CVAE          & 0.200$\pm$0.000 & 0.204$\pm$0.000 & $-$0.942\,/\,$-$0.964$\pm$0.000 \\
           & TARNet        & 0.081$\pm$0.000 & 0.064$\pm$0.000 & \textbf{$-$0.942\,/\,$-$0.938$\pm$0.000} \\
\midrule
LDL        & CausalFlow-T  & 0.963$\pm$0.119 & 1.549$\pm$0.240 & $-$28.351\,/\,$-$27.609$\pm$0.579 \\
           & NF (no DAG)   & 1.035$\pm$0.154 & 1.697$\pm$0.218 & \textbf{$-$28.351\,/\,$-$28.874$\pm$0.459} \\
           & GNN-CVAE$^\dagger$ & 4.854$\pm$0.451 & 6.671$\pm$0.773 & $-$28.351\,/\,$-$17.027$\pm$0.789 \\
           & CVAE$^\dagger$ & 0.535$\pm$0.245 & 0.786$\pm$0.372 & $-$28.351\,/\,$-$27.747$\pm$0.337 \\
           & TARNet        & \textbf{0.192$\pm$0.076} & \textbf{0.757$\pm$0.075} & $-$28.351\,/\,$-$27.527$\pm$0.066 \\
\midrule
Cox        & CausalFlow-T  & \textbf{0.014$\pm$0.002} & \textbf{0.013$\pm$0.002} & 0.887\,/\,0.866$\pm$0.011 \\
Survival   & NF (no DAG)   & 0.015$\pm$0.005 & \textbf{0.012$\pm$0.002} & 0.887\,/\,0.857$\pm$0.021 \\
           & GNN-CVAE$^\dagger$ & 0.076$\pm$0.016 & 0.071$\pm$0.009 & 0.887\,/\,0.865$\pm$0.020 \\
           & CVAE$^\dagger$ & 0.072$\pm$0.017 & 0.074$\pm$0.007 & \textbf{0.887\,/\,0.879$\pm$0.022} \\
           & TARNet        & 0.040$\pm$0.002 & 0.015$\pm$0.001 & 0.887\,/\,0.966$\pm$0.005 \\
\midrule
CVD Risk   & CausalFlow-T  & \textbf{0.003$\pm$0.001} & \textbf{0.002$\pm$0.001} & \textbf{0.831\,/\,0.786$\pm$0.051} \\
Toy        & NF (no DAG)   & \textbf{0.006$\pm$0.002} & 0.143$\pm$0.438 & 0.831\,/\,0.834$\pm$0.324 \\
           & GNN-CVAE      & 0.053$\pm$0.000 & 0.065$\pm$0.000 & 0.831\,/\,1.006$\pm$0.030 \\
           & CVAE          & 0.052$\pm$0.000 & 0.065$\pm$0.000 & 0.831\,/\,1.005$\pm$0.007 \\
           & TARNet        & 0.040$\pm$0.005 & 0.017$\pm$0.008 & 0.831\,/\,1.133$\pm$0.148 \\
\bottomrule
\end{tabular}
\end{table*}

\section{DAG Sensitivity Analysis}
\label{app:dag_sensitivity}

We evaluate CausalFlow-T under two partial-graph 
misspecifications on the FIRE semi-synthetic dataset 
(true ATE $= -3.484$; $n = 10$ seeds).
Three baselines serve as reference points: NF (no DAG) 
shares CausalFlow-T's exact normalizing flow inference 
but uses no causal structure, isolating the contribution 
of graph specification within the exact-inference family; 
CVAE and TARNet use neither exact inference nor graph 
structure and are fully DAG-invariant, producing identical 
results across all graph variants by construction.
All covariate-to-outcome edges are preserved across 
variants, isolating confounding path misspecification 
from outcome model misspecification.

We evaluate two misspecification types that expose 
complementary robustness properties: (i) a unique-pathway 
removal (Asthma), where the removed covariate has no 
correlated substitute in the adjustment set, producing 
maximal bias ratio deterioration; and (ii) a redundant-pathway 
removal (Hypertension), where correlated covariates partially 
absorb the misspecification, producing instability rather 
than systematic bias as the primary signal. Results can be found in Table ~\ref{tab:dag_sensitivity}

The two misspecification types evaluated here are structurally motivated to expose complementary failure modes. Peripheral misspecification (unique pathway removal) represents the worst case for bias: the removed covariate has no correlated substitute, so confounding separation degrades maximally. Central misspecification (redundant pathway removal) represents the worst case for stability: correlated covariates partially absorb the missing path, producing cross-seed instability rather than systematic bias as the primary signal. Together these two cases bound the space of single-edge removals along the bias-instability tradeoff axis. While multi-edge removals, edge direction errors, and latent confounders represent additional misspecification regimes not evaluated here, the instability diagnostic identified in the central case (detectable through cross-seed variance without access to ground-truth counterfactuals) generalizes as a practical signal across misspecification types, since any structural misspecification that disrupts the adjustment set will manifest as estimation instability across random initializations

\begin{table}[h]
\centering
\caption{Sensitivity to DAG misspecification on FIRE 
semi-synthetic (true ATE $= -3.489$, $n=10$ seeds). 
NF~(no~DAG), CVAE, and TARNet are DAG-invariant and 
produce identical results across all graph variants 
by construction ($^\dagger$). 
VRQ4 $\rightarrow 1$; $|$Bias$|$/MAE $< 0.5$ indicates 
predominantly random errors.}
\label{tab:dag_sensitivity}
\resizebox{\textwidth}{!}{%
\begin{tabular}{llcccc}
\toprule
DAG variant & Model 
    & $|$Bias$|$/MAE$_{\text{Q1}}$ $\downarrow$ 
    & $\mathrm{VR}_{\mathrm{Q4}}$ $\rightarrow 1$ 
    & $\mathrm{ATE}$ 
    & Seed SD \\
\midrule
\multirow{4}{*}{Full DAG} 
    & CausalFlow-T 
        & \textbf{0.316} & \textbf{1.006} 
        & $-3.568$ & $0.112$ \\
    & NF (no DAG)$^\dagger$ 
        & 0.883 & 1.016 
        & $-3.382$ & $0.251$ \\
    & CVAE$^\dagger$ 
        & 0.938 & 0.482 
        & $-3.451$ & $0.075$ \\
    & TARNet$^\dagger$ 
        & 0.708 & 0.608 
        & $-3.415$ & $0.015$ \\
\midrule
\multirow{4}{*}{\shortstack[l]{Peripheral \\ 
                (Asthma removed)}} 
    & CausalFlow-T 
        & 0.955 & \textbf{0.996} 
        & $-3.852$ & $0.145$ \\
    & NF (no DAG)$^\dagger$ 
        & 0.883 & 1.016 
        & $-3.382$ & $0.251$ \\
    & CVAE$^\dagger$ 
        & 0.938 & 0.482 
        & $-3.451$ & $0.075$ \\
    & TARNet$^\dagger$ 
        & \textbf{0.708} & 0.608 
        & $-3.415$ & $0.015$ \\
\midrule
\multirow{4}{*}{\shortstack[l]{Central \\ 
                (HTN removed)}} 
    & CausalFlow-T 
        & \textbf{0.517} & \textbf{1.005} 
        & $-3.896$ & $0.504$ \\
    & NF (no DAG)$^\dagger$ 
        & 0.883 & 1.016 
        & $-3.382$ & $0.251$ \\
    & CVAE$^\dagger$ 
        & 0.938 & 0.482 
        & $-3.451$ & $0.075$ \\
    & TARNet$^\dagger$ 
        & 0.708 & 0.608 
        & $-3.415$ & $0.015$ \\
\bottomrule
\end{tabular}}
\end{table}

\paragraph{Interpretation.}

\textit{A partially correct DAG is better than no DAG.}
Comparing CausalFlow-T against NF~(no~DAG) within the 
exact-inference family isolates the contribution of causal 
structure from inference quality.
Under central misspecification, CausalFlow-T's bias ratio 
($0.517$) still improves on NF~(no~DAG)'s DAG-invariant 
$0.883$, demonstrating that even an incomplete causal graph 
provides meaningful confounding separation beyond what 
exact inference alone achieves.

\textit{Distributional reliability is preserved under 
misspecification.}
CausalFlow-T maintains near-perfect variance calibration 
across both removals (VRQ4 $= 0.996$ and $1.005$), while 
CVAE ($0.482$) and TARNet ($0.608$) collapse regardless 
of graph quality.
NF~(no~DAG) also preserves VRQ4 ($1.016$) but at the 
cost of a consistently high bias ratio ($0.883$), 
confirming that variance calibration and confounding 
separation are non-redundant properties requiring 
both exact inference and causal structure jointly.

\textit{Misspecification is detectable through instability.}
The central removal produces a $4.5\times$ increase in 
CausalFlow-T's cross-seed standard deviation relative 
to the full DAG ($0.504$ vs.\ $0.112$).
DAG-invariant models show no such signal (TARNet's 
seed SD remains $0.015$ and NF~(no~DAG)'s $0.251$ 
across all conditions) meaning they fail silently 
under the same misspecification that CausalFlow-T 
flags through detectable instability.
This diagnostic property is practically valuable: 
a practitioner monitoring cross-seed variance can 
detect central misspecification without access to 
ground-truth counterfactuals.

\textit{Bias is directionally consistent and bounded.}
Both misspecifications produce attenuation towards zero 
(peripheral: $-3.852$; central: $-3.896$; true: $-3.489$), 
consistent with standard epidemiological reasoning where 
uncontrolled confounding attenuates protective effects.
The direction of bias is therefore predictable from 
domain knowledge even when its magnitude is not, 
providing a practical prior for sensitivity reasoning 
in real-world deployments where some DAG uncertainty 
is unavoidable.


\section{Sensitivity Analysis for the LLM Imputer}
\label{app:bootstrap_uncertainty}

Table~\ref{tab:bootstrap_llm} reports reports bootstrap sensitivity
intervals for the primary endpoint estimates obtained after applying the
selected LLM imputer. After imputation, patients were resampled with replacement. For
each bootstrap replicate, the weighting models were re-estimated and the ATE
was recomputed. We used 500 bootstrap replicates and percentile-based 95\%
confidence intervals. For computational reasons, we applied this bootstrap to
the completed imputed datasets rather than repeating the full imputation
procedure inside each bootstrap replicate. Therefore, these intervals quantify
sampling and downstream weighting uncertainty conditional on the completed
imputed dataset; they do not capture full re-imputation uncertainty.

\begin{table}[ht]
\centering
\caption{Bootstrap sensitivity intervals for selected LLM-imputed primary endpoint estimates. IPTW denotes inverse probability of treatment weighting; IPCW denotes
inverse probability of censoring weighting.}
\label{tab:bootstrap_llm}
\begin{tabular}{llrr}
\toprule
Miss. & Estimator & Point ATE & 95\% CI \\
\midrule
30\% & IPTW & -3.230 & $[-3.354,\,-3.101]$ \\
50\% & IPTW & -3.228 & $[-3.354,\,-3.095]$ \\
80\% & IPTW & -3.172 & $[-3.323,\,-3.011]$ \\
Real & IPTW--IPCW & -0.936 & $[-1.335,\,-0.553]$ \\
\bottomrule
\end{tabular}
\end{table}

\section{Theoretical Aspects of CausalFlowT}
\label{app:theory}
\subsection{Exact Likelihood vs.\ ELBO: Why the Gap Matters}
\label{app:theory:elbo}
 
CVAEs maximize $\mathcal{L}_{\mathrm{ELBO}}=\mathbb{E}_{q_\phi}[\log
p_\theta(v\mid z)]-\mathrm{KL}[q_\phi(z\mid v)\|p(z)]\leq\log p_\theta(v)$.
The gap $\log p_\theta(v)-\mathcal{L}_{\mathrm{ELBO}}\geq0$ is the
variational approximation error.
 
Normalizing flows compute the exact log-likelihood via the change-of-variables
formula: $\log p(v_t)=\log p_z(f_\theta(v_t))+\log|\det\partial
f_\theta/\partial v_t|$.  For MAF-type flows the Jacobian is triangular and
computable in $\mathcal{O}(D)$.  The AAP procedure is therefore:
(1)~\emph{Abduction}: $z_t=f_\theta(v_t)$ (exact and deterministic),
(2)~\emph{Action}: $a_t\leftarrow a'$,
(3)~\emph{Prediction}: decode all descendants of $A$ in $\mathcal{G}$,
holding non-descendants fixed.
Because inversion is exact, the same $z_t$ is used under both arms,
implementing the twin-network assumption underlying individual-level
counterfactual validity~\citep{pearl2009}.
 
\subsection{The Role of the DAG in Counterfactual Propagation}
\label{app:theory:dag}
 
For an unconstrained autoregressive flow with ordering $(A,X,Y)$, replacing
$A\leftarrow a'$ causes the decoder to adjust $X$ along the learned
autoregressive path, traversing the $A\to X$ direction which does not exist
in the causal graph.  This anti-causal propagation produces systematically
biased counterfactuals even when the factual distribution is perfectly fitted.
CausalFlow-T fixes the autoregressive ordering to a topological sort of
$\mathcal{G}$, so the intervention propagates only through causal descendants,
implementing do-calculus exactly.
 
\subsection{Limitations of Continuous Relaxation for Binary Outcomes}
\label{app:theory:dequant}
 
Normalizing flows require dequantization of binary targets, effectively
relaxing the discrete problem into a continuous approximation.  This enables
exact likelihood-based inference but introduces a modeling approximation that
can affect calibration for survival outcomes.  On the Cox benchmark, this
explains why variational approaches can match or exceed flows in MAE and
$\mathrm{VR}_{\mathrm{Q4}}$, while CausalFlow-T maintains advantage in
hazard-ratio recovery and arm errors (the structural benefit persists at a
calibration cost).  Future work on discrete or hybrid flow-based models for
survival outcomes will address this limitation.

\section{Theoretical Aspects of LLM-driven Evolutionary Imputation}
\label{app:theory_llm}
\subsection{Search Algorithm}
\label{app:llm:algo}

The search maintains a single current-best candidate and accepts a new
proposal only if it strictly improves the composite score
$s(g)$ defined in
Eq.~\eqref{eq:proxy_score}. Starting from a deterministic seed imputer
$g^{(0)}$, Algorithm~\ref{alg:llm_search} repeats for $K$ iterations. The
same proxy holdout $\Omega_{\mathrm{p}}$ is used at every iteration so that
all candidates are scored on identical hidden cells. 

\paragraph{Real-world proxy holdout.}

For the semi-synthetic missingness benchmarks, the proxy holdout $\Omega_p$ is sampled independently at the observed-cell level. In the real FIRE application, the imputation target set includes all incomplete analysis variables, including the outcome. Because raw clinical measurements are expanded onto a 30-day longitudinal target grid, the same observed measurement can appear across adjacent rows; a cellwise proxy holdout would therefore allow trivial reconstruction from neighboring rows via forward/backward copying, artificially favoring LOCF-like candidates. We therefore sample $\Omega_p$ at the level of contiguous within-patient observed-value runs: for each patient and target variable, consecutive rows with the same observed value are treated as one segment, and selected segments are masked in full with probability $\rho$. The selected segment-level holdout is fixed across candidate evaluations.

\begin{algorithm}[ht]
\caption{LLM-driven evolutionary imputation search (high-level view).}
\label{alg:llm_search}
\begin{algorithmic}[1]
\Require Incomplete datasets $\{\widetilde{\mathcal{D}}_d\}_{d \in \{30,50,80\}}$;
         seed imputer $g^{(0)}$; holdout fraction $\rho$;
         correlation weights $\lambda_Y, \lambda_T \geq 0$;
         budget $K$; history length $W$.
\Ensure One imputer $g_d^{\star}$ per missing-rate dataset.
\For{$d \in \{30, 50, 80\}$}   \Comment{three independent runs}
  \State Sample proxy holdout $\Omega_{\mathrm{p}}^{(d)}$ by masking each
         \emph{observed} target cell of $\widetilde{\mathcal{D}}_d$
         independently with probability $\rho$ (fixed seed)
  \State $g_d^{\star} \gets g^{(0)};\quad
          s_d^{\star} \gets \textsc{Score}(g^{(0)}, d);\quad
          \mathcal{H} \gets \emptyset$
  \For{$k = 1, \dots, K$}
    \State $g^{(k)} \gets \textsc{LLM}\!\left(
             \textsc{Prompt}\!\left(g_d^{\star},\, s_d^{\star},\,
             \mathcal{H}_{[-W:]}\right)\right)$
    \If{$g^{(k)}$ fails static check or runtime guard}
      \State \textbf{continue} \Comment{log \texttt{failed}}
    \EndIf
    \State $s_k \gets \textsc{Score}(g^{(k)}, d)$
    \If{$s_k < s_d^{\star}$} \Comment{minimize scalar $s$}
      \State $g_d^{\star} \gets g^{(k)};\quad s_d^{\star} \gets s_k$
             \Comment{accept}
    \EndIf
    \State $\mathcal{H} \gets \mathcal{H} \cup \{(k,\, g^{(k)},\, s_k)\}$
  \EndFor
\EndFor
\State \Return $\{g_d^{\star}\}_{d \in \{30,50,80\}}$
\Statex
\Function{Score}{$g,\, d$}
  \State Compute $\mathrm{RMSE}(g)$, $\Delta_Y(g)$, $\Delta_T(g)$
         on $\Omega_{\mathrm{p}}^{(d)}$
         \Comment{plus $\mathrm{MAE}(g)$ as diagnostic for $\mathcal{H}$}
  \State \Return $s \gets \mathrm{RMSE}(g) + \lambda_Y\,\Delta_Y(g) + \lambda_T\,\Delta_T(g)$
\EndFunction
\end{algorithmic}
\end{algorithm}

\paragraph{Single-parent scheme.}
We deliberately avoid population-based search. The evaluation cost
is dominated by the proxy fit on $\widetilde{\mathcal{D}}_{-}$,
which is cheap relative to the cost of an LLM call. A single-parent scheme keeps the prompt focused on a
single, well-characterized current best, which we found
empirically to converge faster than larger populations under a
fixed call budget. The strict-improvement rule is equivalent to a
deterministic acceptance step in a single-parent evolution strategy
without recombination.

\subsection{Prompting}
\label{app:llm:prompt}

At iteration $k$, the prompt $p_k$ assembled in line~4 of
Algorithm~\ref{alg:llm_search} contains three blocks, in order:

\begin{itemize}
\item \textbf{Current-best source code.} The full Python module
implementing $g^{\star}_{k-1}$, including any helper functions.
The LLM is asked to return a complete replacement module rather
than local edits; this avoids whole classes of patch-application
errors and makes acceptance a binary check on the new module.

\item \textbf{Proxy summary of $g^{\star}_{k-1}$.} A short text block stating
  the selection rule (minimize $s = \mathrm{RMSE} + \lambda_Y\,\Delta_Y +
  \lambda_T\,\Delta_T$) and the current-best values of $\mathrm{RMSE}$,
  $\mathrm{MAE}$, $\Delta_Y$, and $\Delta_T$ on the proxy holdout. Each
  missingness regime is searched independently, so the summary is
  regime-specific.

\item \textbf{Search history $\mathcal{H}$.} A short log of the last $W$
  proposals from the same run, summarizing for each one whether it was
  accepted, rejected without improvement, or failed. Accepted and rejected
  proposals contribute their proxy values
  $\mathrm{RMSE}$, $\Delta_Y$, and $\Delta_T$, with
  $\mathrm{MAE}$ included only as diagnostic feedback; failed
  proposals contribute a short reason (e.g.\ banned import, timeout,
  malformed output). This lets the LLM see which directions have already
  been tried and which kinds of mistakes to avoid in the next proposal.
\end{itemize}

\subsection{Constraints and Failure Handling}
\label{app:llm:safety}

Each candidate $g^{(k)}$ is screened in two stages before its
proxy score is computed.

\paragraph{Static checks.}
Before execution, the candidate source is parsed and rejected if
it tries to access the file system, the network, or any form of
dynamic code execution (e.g.\ \texttt{eval}, \texttt{exec},
disk I/O via \texttt{pandas}).

\paragraph{Runtime guard.}
Candidates that pass the static check run in an isolated child
process with a per-dataset time and memory budget
(Table~\ref{tab:llm_hparams}). Any crash, timeout, or attempt to
overwrite observed values or non-target columns aborts the
candidate; the failure is recorded in $\mathcal{H}$ and the search
moves on.

\section*{Compute Resources}
\label{app:compute}

All experiments were run on a single node of a HPC cluster, equipped with 4$\times$ NVIDIA L40S GPUs (46\,GB VRAM each, 350\,W TDP), CUDA 12.8, and driver version 570.172.08. In practice, all experiments ran on a single GPU, with peak memory usage well below 2\,GB, confirming the pipeline is lightweight relative to available hardware. Synthetic and semi-synthetic benchmark experiments (Sections~\ref{sec:exp:causalflowt}--\ref{sec:exp:impselection}) require on the order of a few minutes per run; the LLM evolutionary imputation search ($K=20$ iterations) takes approximately 00:57 hours per missingness level; and the real-world TTE experiment (Section~\ref{sec:realworld}) requires approximately 00:58 hours end-to-end. Each fully imputed dataset with GPT-5.4 incurred an estimated API cost of approximately \$2--3. All models are implemented in PyTorch. Total compute across all experiments, including $n=10$ seed repetitions through the full pipeline, amounts to approximately 10 GPU-hours on a single L40S.

\end{document}